\documentclass[lettersize,journal]{IEEEtran}
\usepackage{amsmath,amsfonts}
\usepackage{algorithmic}
\usepackage{algorithm}
\usepackage{array}
\usepackage[caption=false,font=normalsize,labelfont=sf,textfont=sf]{subfig}
\usepackage{textcomp}
\usepackage{stfloats}
\usepackage{url}
\usepackage{verbatim}
\usepackage{graphicx}
\usepackage{cite}
\usepackage{booktabs}
\usepackage{dsfont}
\usepackage{pifont}
\usepackage{multirow}
\usepackage{multicol}
\usepackage{threeparttable}
\usepackage[table]{xcolor}
\usepackage{tabularx}
\usepackage{arydshln}
\definecolor{revisionred}{RGB}{180,30,30}
\definecolor{lightbg}{HTML}{F2F6F9}
\definecolor{carhead}{HTML}{E6F0FA}
\definecolor{cubhead}{HTML}{FFF4CC}
\definecolor{airhead}{HTML}{E6F4EA}

\begin{document}

\title{Attribute-Evidence Learning with Multimodal Prompt for Generalized Category Discovery}
%From Localized Attribute Evidence to Global Visual Reasoning: A Multimodal Prompt Transformer for Generalized Category Discovery
%Taming Textual Guidance via Decoupled Multimodal Learning for Generalized Category Discovery
% Decoupled Textual Guidance for Multimodal Generalized Category Discovery
% Decoupled Multimodal Prompt Learning for Generalized Category Discovery
% Taming Textual Guidance for Multimodal Generalized Category Discovery

\author{Wei Zhang, Zhu Teng, Zhixiang Chi, Junnan Zou, Wenxin Luo, Linfeng Ye, Jianping Fan, and Konstantions N. Plataniotis,~\IEEEmembership{Fellow,~IEEE,}
        % <-this % stops a space
\thanks{Wei Zhang, Zhu Teng(Corresponding Author), Junnan Zou and Wenxi Luo are with the Beijing Jiaotong University, Beijing 100044, China. (E-mail: {wzhang0926, zteng, zoujunnan, wxluo0903}@bjtu.edu.cn)}% <-this % stops a space
\thanks{Zhixiang Chi, Linfeng Ye and Konstantions N. Plataniotis are with University of Toronto, Toronto, ON M5S 2E8, Canada. (E-mail: {zhxchi, kostas}@ece.utoronto.ca, linfeng.ye@mail.utoronto.ca)}
\thanks{Jianping Fan is with the FuDan University, Shanghai 200433, China. (E-mail: jianpingfan@fudan.edu.cn)}

}

\markboth{Journal of \LaTeX\ Class Files,~Vol.~14, No.~8, August~2021}%
{Shell \MakeLowercase{\textit{et al.}}: A Sample Article Using IEEEtran.cls for IEEE Journals}


\maketitle

\begin{abstract}
Generalized category discovery (GCD) aims to discover novel categories from unlabeled data by leveraging limited labeled samples from old classes. Existing methods that rely solely on visual cues often neglect the multimodal nature of human perception, prompting recent approaches to incorporate textual information such as category tags and attributes.
However, these approaches typically entangle global tags and local attributes within a shared global image representation, rendering attributes unreliable and potentially providing misleading cues that can drive the model toward incorrect predictions.
To this end, we propose AptM-GCD, a novel multimodal framework that exploits apt textual guidance through attribute evidence learning and multimodal prompt reasoning. Specifically, we design an Attribute-Evidence Learning (AEL) mechanism that treats attributes as localized visual evidence rather than global category descriptors. By interacting attribute embeddings with patch-level visual tokens, AEL identifies reliable attribute-related regions and extracts discriminative local evidence, thereby reducing the impact of noisy or irrelevant attribute cues. Building upon the extracted evidence, we further develop a Multimodal Prompt Learning (MPL) mechanism to bridge local attribute semantics and global visual understanding. MPL converts fine-grained textual evidence into visual prompts and injects them into a multimodal prompt transformer, enabling attribute information to enhance category discrimination through contextual reasoning without enforcing direct global alignment between modalities.
Extensive experiments on six datasets demonstrate that AptM-GCD consistently achieves SoTA performance across both fine-grained and generic benchmarks. Moreover, the proposed framework exhibits strong generalization ability, delivering consistent gains under unimodal and class-imbalanced GCD settings.
\end{abstract}

\begin{IEEEkeywords}
Generalized Category Discovery, Multimodal Representation, Semi-Supervised Learning, Prompt Learning.
\end{IEEEkeywords}
\section{Introduction}
\label{sec:intro}

\IEEEPARstart{H}{umans} possess a remarkable ability to recognize and organize novel concepts from limited prior experience, continuously expanding their knowledge of the world \cite{humannovel1,humannovel2}. Equipping deep learning models with a similar capability is crucial for open-world visual understanding, as real-world data distributions are constantly evolving and often contain categories that are absent from the labeled training set \cite{deeplearning1,deeplearning2,deeplearning3}. Generalized Category Discovery (GCD) has recently emerged as a promising framework for addressing this challenge. Given a small labeled dataset consisting of known categories and a large unlabeled dataset containing both known and unknown categories, GCD aims to simultaneously preserve recognition of known classes while discovering the underlying structure of novel categories \cite{gcd,dccl,simgcd}. This setting better reflects realistic learning scenarios, where labeled annotations are limited and new semantic concepts continuously appear. Compared with related paradigms, GCD poses a substantially more challenging problem. Novel Category Discovery (NCD) assumes that all unlabeled samples belong exclusively to unseen classes \cite{ncd1,ncd2,ncd3}, while Open-Set Recognition (OSR) focuses primarily on identifying whether a sample is novel without further categorizing it \cite{osr1,osr2}. In contrast, GCD must not only distinguish known and unknown classes but also group unlabeled samples into semantically meaningful novel categories, requiring the model to balance representation transfer, class discrimination, and category discovery within a unified framework.

\begin{figure}[t]
\centering
\includegraphics[width=\linewidth]{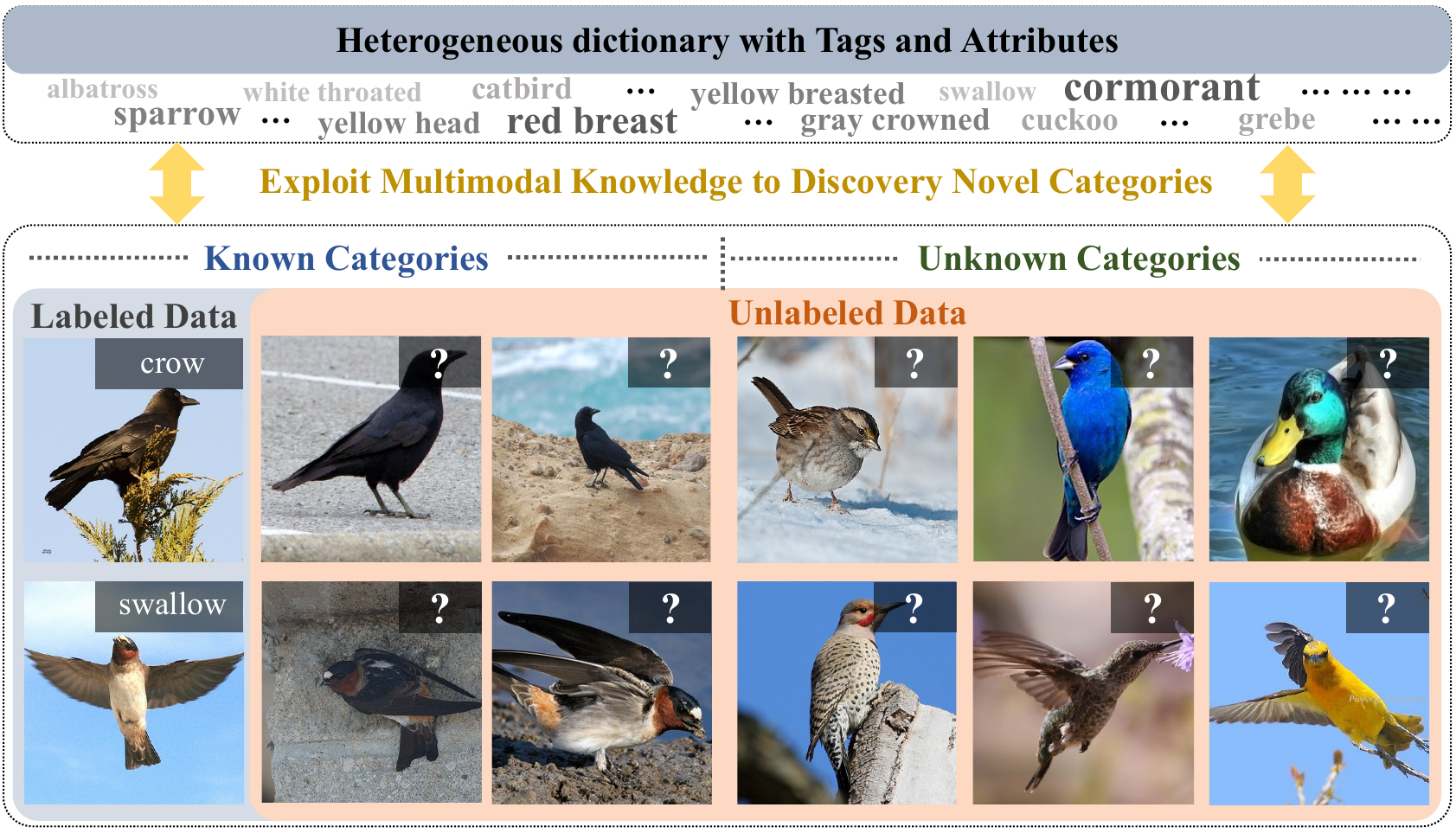}
\caption{\textbf{Illustration of multimodal generalized category discovery. }Images include labeled data from old classes and unlabeled data from both old and novel categories, while a heterogeneous textual dictionary of category tags and attribute descriptions introduces semantic guidance for category discovery.}
\label{fig:first}
\vspace{-10pt}
\end{figure}

Early GCD methods primarily rely on visual representations to discover novel categories. However, human perception naturally integrates visual observations with semantic descriptions and fine-grained attributes when recognizing unfamiliar concepts \cite{humanmultimodal}. Inspired by this observation, recent studies have incorporated textual knowledge into GCD \cite{clipgcd}, leveraging category tags and attribute descriptions to provide complementary semantic supervision and enhance representation learning \cite{textgcd,ssr2gcd}.
A common strategy in these multimodal approaches is to construct a heterogeneous vocabulary that combines category-level concepts (e.g., \textit{sparrow}) with attribute-level descriptions (e.g., \textit{gray crown}), as illustrated in Fig.~\ref{fig:first}. Given a query image, the model retrieves semantically relevant textual items through global image-level matching and subsequently converts the retrieved tags and attributes into pseudo labels for cross-modal co-teaching with visual category predictions \cite{textgcd,ssr2gcd}. Such a design implicitly treats category tags and attributes as equally informative global semantic descriptors, allowing textual information to directly influence category discovery.

\begin{figure*}[t]
\centering
\includegraphics[width=\linewidth]{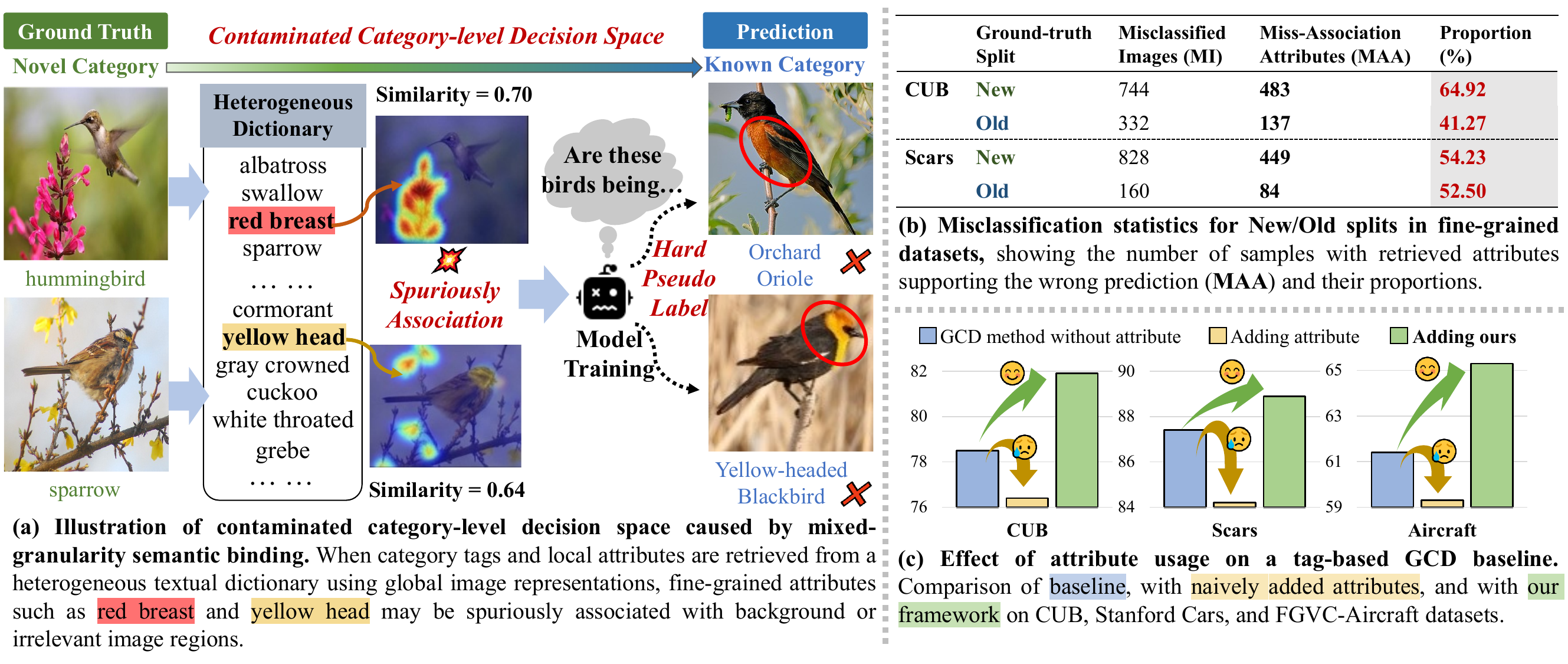}
\caption{\textbf{Overview of challenges in existing multimodal GCD methods.} \textbf{(a) Attribute-based GCD pipeline: }Illustration of how mixed-granularity semantic binding can cause spurious attribute-image associations, contaminating the category-level decision space. \textbf{(b) Empirical misclassification statistics:} For attribute-based methods \cite{textgcd}, showing the proportion of samples where retrieved attributes are irrelevant to the ground-truth class and lead to incorrect predictions. \textbf{(c) Direct addition of attributes to tag-based GCD:} Naively adding attributes to a tag-based method can degrade performance on novel classes, whereas our proposed method improves results.}
\label{fig:intro}
\vspace{-10pt}
\end{figure*}
Although these methods introduce rich semantic guidance, they overlook a fundamental issue: \textit{\textbf{attributes and category tags operate at different semantic granularities}}. Category tags describe holistic class identity and serve as image-level semantic anchors, whereas attributes correspond to localized visual evidence, such as color, texture, shape, or part-level patterns. Furthermore, the validity of an attribute often depends on the object category. For instance, the attribute \textit{blue} may appear in many contexts, but it is not a reliable descriptor for a \textit{tomato} in a natural scene. When performing global image-level retrieval, the attribute \textit{blue} might still be selected if the background contains blue regions. As illustrated by the examples in Fig.~\ref{fig:intro}(a), such retrieval can mistakenly associate local attributes with irrelevant background or contextual cues. Consequently, the retrieved attributes may provide misleading yet seemingly plausible semantic evidence, which can steer the model toward incorrect category predictions.

This issue is particularly detrimental in the context of GCD, where novel classes lack labeled supervision and must be inferred solely from unlabeled data. As a result, the model heavily depends on the semantic structure provided by external textual guidance. When noisy or granularity-mismatched attributes are retrieved, they can bias the text guidance toward incorrect category-level predictions. These biased predictions are subsequently transformed into pseudo labels and propagated through cross-modal co-teaching. Consequently, erroneous attribute evidence contaminates the category discovery space, impairing the reliable separation of novel classes.

To quantify this effect, Fig.~\ref{fig:intro}(b) reports the attribute-induced prediction bias on fine-grained CUB \cite{cub} and Stanford Cars \cite{cars}, using the attribute-based method \cite{textgcd}. Among misclassified novel-class samples, 64.92\% on CUB and 54.23\% on Stanford Cars involve retrieved attributes that provide misleading support for the predicted category, highlighting the substantial impact of noisy attributes on GCD performance.

Beyond attribute-based GCD methods, this finding also raises an important question for tag-based approaches: can attributes simply be added as extra textual guidance?
Fig. \ref{fig:intro}(c) shows that the answer is not straightforward.
Naively incorporating attributes into a tag-based GCD baseline \cite{spectralgcd} does not consistently improve novel-class discovery and may even degrade performance.
In contrast, when attributes are decoupled from category tags and integrated through our proposed strategy, they yield clear improvements over both the original baseline and naive attribute integration.
These observations indicate that rich attribute semantics are beneficial only when their granularity is properly handled, rather than being treated as additional global category-level textual anchors.

To this end, we propose AptM-GCD, a multimodal GCD framework that handles textual semantics by granularity: fine-grained attributes serve as localized visual evidence to refine image representations, while category tags act as global semantic anchors for final category discovery.
Specifically, we introduce Attribute-Evidence Learning (AEL), an object-aware attribute grounding mechanism. Rather than retrieving attributes from a global image representation, AEL performs attribute mining over patch-level visual tokens conditioned on object regions. This allows attributes to be selected and grounded according to object-relevant visual evidence, suppressing noisy cues from background or irrelevant context.
Building upon grounded evidence, we further propose Multimodal Prompt Learning (MPL). MPL transforms localized attribute evidence into soft visual prompts and injects them into a transformer-based visual backbone. This enables fine-grained attribute semantics to interact with global visual context, refining image representations without forcing attributes into global category-level alignment. Beyond AptM-GCD, MPL is architecture-agnostic and can be applied to both multimodal and unimodal GCD frameworks to improve category discovery without altering original learning objectives (Sec.~\ref{sec:exp-mpl}). To summarize, our contributions are four-fold:

\begin{itemize}
    \item We identify a previously overlooked granularity mismatch in multimodal GCD, where local attributes are treated as global category-level semantics, potentially biasing novel-category discovery. To address this issue, we propose AptM-GCD, which leverages attributes as grounded visual evidence rather than category-level supervision.

    \item We propose Attribute-Evidence Learning (AEL), which performs object-conditioned attribute mining and attribute-to-patch grounding, replacing global image-level attribute alignment with localized evidence extraction.

    \item We design Multimodal Prompt Learning (MPL), which encodes localized attribute evidence as soft visual prompts and injects them into the visual transformer, facilitating interaction between fine-grained attribute cues and global image context. MPL is architecture-agnostic and can also be seamlessly integrated into existing multimodal GCD methods without altering their architectures.

    \item Extensive experiments demonstrate that AptM-GCD achieves state-of-the-art performance across multiple benchmarks. Beyond multimodal GCD, MPL also shows strong generalization to unimodal and unbalanced GCD settings, further validating its flexibility and effectiveness.
\end{itemize}

% 需要补充和cvpr25那篇论文的区别
% This work is an extension of our previous CVPR 2025 study, AptGCD \cite{aptgcd}, which introduced a vision prompt learning method to address the scattered-attention in unimodal GCD. The statement describing the substantial extensions over the conference version, together with a copy of the prior publication, has been provided in the submission materials.

% Compared with AptGCD, this extension differs in three major aspects. 

% \textbf{(1) Problem formulation: }we move beyond unimodal representation learning and investigate a new challenge in multimodal GCD: fine-grained attributes, when naively aligned with global image representations together with category tags, can provide misleading textual guidance and bias novel-category discovery. 

% \textbf{(2) Methodology:} we propose AptM-GCD, a novel multimodal framework that disentangles category-level tags from local attribute cues, introduces an Attribute-Evidence Learning mechanism for patch-grounded attribute learning, and develops Multimodal Prompt Learning to inject localized attribute evidence into global visual reasoning.

% \textbf{(3) Experiments:} we provide new empirical analyses of attribute-induced misclassification and conduct extensive evaluations across multimodal, unimodal, fine-grained, generic, and class-imbalanced GCD settings, demonstrating the broader effectiveness and generality of the proposed framework.
\section{Related Work}
\label{sec:rl}
% \paragraph{\textbf{Generalized Category Discovery}}
% \textbf{Generalized Category Discovery}
% is first introduced by \cite{gcd} as an extension of NCD. It fine-tunes target-domain representations via contrastive learning and then applies semi-supervised k-means to group unlabeled samples from both seen and novel categories. Subsequent studies have mainly improved the representation learning and clustering stages. DCCL \cite{dccl} introduces a two-level contrastive objective to capture inter-class semantic relations. PromptCAL \cite{promptcal} alleviates false-negative issues by mining reliable positives from affinity graphs, while CiPR \cite{cipr} adopts a stronger contrastive scheme together with semi-supervised hierarchical clustering. CMS \cite{cms} introduces the mean-shift-based contrastive learning. Besides, SimGCD \cite{simgcd} replaces explicit clustering with a prototype classifier for pseudo-label assignment. Based on SimGCD, LegoGCD \cite{legogcd} introduces entropy regularization to mitigate catastrophic forgetting. AllGCD \cite{allgcd} incorporates all unlabeled samples into contrastive learning, and ProtoGCD \cite{protogcd} jointly models old and novel classes with shared learnable prototypes.

\textbf{Generalized Category Discovery} was first introduced by \cite{gcd} as an extension of NCD, combining contrastive representation learning with semi-supervised clustering to discover both old and novel categories. Subsequent studies have mainly focused on improving representation learning and clustering strategies. For instance, DCCL \cite{dccl} captures inter-class relations with a two-level contrastive objective, PromptCAL \cite{promptcal} mines reliable positives to alleviate false negatives, CiPR \cite{cipr} enhances contrastive learning with hierarchical clustering. 
Moreover, recent multi-view clustering studies have shown that reliable cluster discovery can be further improved by exploiting complementary structures across different views, \textit{e.g.,} through selective cross-view topology learning \cite{dong1} and cross-view graph matching\cite{dong2}. Different from these clustering-based paradigms, SimGCD \cite{simgcd} replaces explicit clustering with a prototype classifier. Other works, such as LegoGCD \cite{legogcd}, AllGCD \cite{allgcd}, and ProtoGCD \cite{protogcd}, further improve pseudo-label assignment, mitigate forgetting, or jointly model old and novel classes with shared prototypes.

% \subsection{Multimodal Generalized Category Discovery}
% \paragraph{\textbf{Multimodal Generalized Category Discovery.}}
\textbf{Multimodal Generalized Category Discovery.}
While early GCD methods rely on visual representations, recent studies have explored incorporating textual information to enhance category discovery. CLIP-GCD \cite{clipgcd} is the first to leverage CLIP representations, combining image and text features to improve clustering quality. Building upon this direction, tag-based methods further exploit image-text supervision: GET \cite{get} synthesizes pseudo textual embeddings from image features and jointly optimizes visual and textual branches with cross-modal consistency, while SpectralGCD \cite{spectralgcd} learns unified cross-modal representations by modeling images as semantic mixtures.
More recently, TextGCD \cite{textgcd} and SSR$^2$-GCD \cite{ssr2gcd} introduce attribute-aware textual guidance. These methods typically adopt a global retrieval paradigm, retrieving tags and attributes from image-level representations and converting them into pseudo labels for novel-category discovery. However, by aligning local attributes with global category tags, they overlook the granularity mismatch between fine-grained attributes and image-level semantics, which may introduce noisy supervision and hinder the discovery of novel categories.

% \subsection{Multimodal Prompt Learning}

\textbf{Prompt Learning in Vision-Language Models.}
Prompt learning has emerged as an efficient strategy for adapting vision-language models (VLMs) to downstream recognition tasks without fully fine-tuning large-scale pretrained models. Existing approaches, such as CoOp \cite{coop}, CoCoOp \cite{cocoop}, VPT \cite{vpt}, MaPLe \cite{maple}, and PromptSRC \cite{promptsrc}, mainly explore learnable prompts in either the textual, visual, or joint vision-language space to improve task adaptation.
Recent works, such as ArGue \cite{argue} and ATPrompt \cite{atprompt}, incorporate attribute descriptions into prompting to enrich fine-grained textual representations. However, these methods typically treat attributes as textual or visual prompts without explicit grounding in local visual regions, which may limit their effectiveness in modeling fine-grained visual evidence. 
%Different from these methods, our MPL follows a learning-to-learn prompting strategy rather than directly optimizing textual or visual prompts. Instead of using attributes as textual prompts, MPL grounds them into localized visual evidence and are injected into the visual transformer to link fine-grained attribute cues with global visual context for representation refinement.

% Prompt learning efficiently adapts vision-language models (VLMs) to downstream recognition tasks, including textual prompting methods such as CoOp \cite{coop} and CoCoOp \cite{cocoop}, and visual prompt tuning methods such as VPT \cite{vpt}. Beyond unimodal prompting, multimodal methods jointly adapt visual and textual representations, e.g., MaPLe \cite{maple} learns coupled vision-language prompts and PromptSRC \cite{promptsrc} regularizes prompt learning to preserve VLM generalization. 

% Recent works further incorporate attribute knowledge into prompting to enrich fine-grained semantic representations.
% ArGue \cite{argue} leverages attribute descriptions to guide prompt tuning, TAP \cite{tap} organizes attributes into structured concept-attribute-description prompts, and ATPrompt \cite{atprompt} introduces anchored attributes to enhance textual prompting for unknown categories. 
\section{Preliminaries}
\label{sec:pre}
\subsection{Problem Formulation of GCD}

\textbf{Conventional GCD}. GCD frames learning as an open-world problem by splitting the dataset into a labeled set $\mathcal{D}_l=\left\{\left(X_i, y_i\right)\right\}_{i=1}^{N_l} \in \mathcal{X}_l \times \mathcal{Y}_l$ and an unlabeled set $\mathcal{D}_u=\left\{\left(X_i, y_i\right)\right\}_{i=1}^{N_u} \in \mathcal{X}_u \times \mathcal{Y}_u$. The classes in $\mathcal{D}_l$, denoted as $\mathcal{Y}_l$, are referred to as old classes, while $\mathcal{D}_u$ contains samples from both old classes and additional novel classes (i.e., $\mathcal{Y}_l \in \mathcal{Y}_u$). The objective of GCD is to exploit the supervision provided by $\mathcal{D}_l$ to recognize samples in $\mathcal{D}_u$ across both known and unknown categories. This requires the model not only to discover novel classes and organize them into meaningful clusters, but also to preserve its ability to distinguish the old classes.

\textbf{Multimodal GCD} extends conventional GCD by introducing additional textual semantics, such as category tags \cite{spectralgcd} and attributes \cite{textgcd, ssr2gcd}. Although the textual vocabulary may contain novel-category names, no visual annotations from novel categories are accessible during training. Hence, textual semantics do not remove the core challenge of GCD, as the model must still learn reliable visual-semantic correspondences for novel categories from unlabeled data. Our work follows this setting and focuses on how such textual knowledge can be effectively exploited for category discovery.

\subsection{Contrastive and Parametric Classification Losses.}
\label{sec:gcd_losses}

Following SimGCD \cite{simgcd}, many recent unimodal and multimodal GCD methods \cite{textgcd,spectralgcd,legogcd,sptnet,protogcd}, including ours, adopt a training framework that combines contrastive representation learning with parametric classification. Given a mini-batch $B$, each image is randomly augmented twice to produce two views, denoted by $\boldsymbol{x}_i$ and $\boldsymbol{x}_i^{\prime}$. A feature extractor $F$ and a projection head $PH$ are used to obtain the contrastive representations
$\boldsymbol{z}_i = PH(F(\boldsymbol{x}_i)).$
The contrastive objective consists of an unsupervised loss over all samples and a supervised loss over labeled samples.
For all data in $B$, the unsupervised contrastive loss encourages two augmented views of the same image to be close while pushing apart different samples:
\begin{equation}
\label{rep1}
\mathcal{L}_{\mathrm{cons}}^{u}
=
\frac{1}{|B|}
\sum_{i\in B}
-\log
\frac{
\exp \left(\boldsymbol{z}_i^\top \boldsymbol{z}_i^{\prime}/\tau_u\right)
}{
\sum_{n\in B, n\neq i}
\exp \left(\boldsymbol{z}_i^\top \boldsymbol{z}_n^{\prime}/\tau_u\right)
}
\end{equation}
where $\tau_u=0.07$ is the temperature parameter. For labeled samples $B^l$, supervised contrastive learning further exploits ground-truth labels by treating samples from the same class as positives. Let $\mathcal{N}_i$ denote the set of samples in the mini-batch that share the same label as $\boldsymbol{x}_i$. The supervised contrastive loss is defined as
\begin{equation}
\label{rep2}
\mathcal{L}_{\mathrm{cons}}^{s}
=
\frac{1}{|B^l|}
\sum_{i\in B^l}
\frac{1}{|\mathcal{N}_i|}
\sum_{q\in \mathcal{N}_i}
-\log
\frac{
\exp \left(\boldsymbol{z}_i^\top \boldsymbol{z}_q^{\prime}/\tau_c\right)
}{
\sum_{n\in B, n\neq i}
\exp \left(\boldsymbol{z}_i^\top \boldsymbol{z}_n^{\prime}/\tau_c\right)
}
\end{equation}
% where $\tau_c=1.0$. The overall constractive learning objective is
% \begin{equation}
% \label{rep3}
% \mathcal{L}_{\mathrm{cons}}
% =
% (1-\lambda)\mathcal{L}_{\mathrm{cons}}^{u}
% +
% \lambda \mathcal{L}_{\mathrm{cons}}^{s},
% \end{equation}
The overall representation learning loss is adjusted using the coefficient $\lambda = 0.35$:
\begin{equation}\label{rep3}
\mathcal{L}_{\text {rep }}=(1-\lambda) \mathcal{L}_{\text {rep }}^u+\lambda \mathcal{L}_{\text {rep }}^s
\end{equation}

\begin{figure*}[t]
\centering
\includegraphics[width=\linewidth]{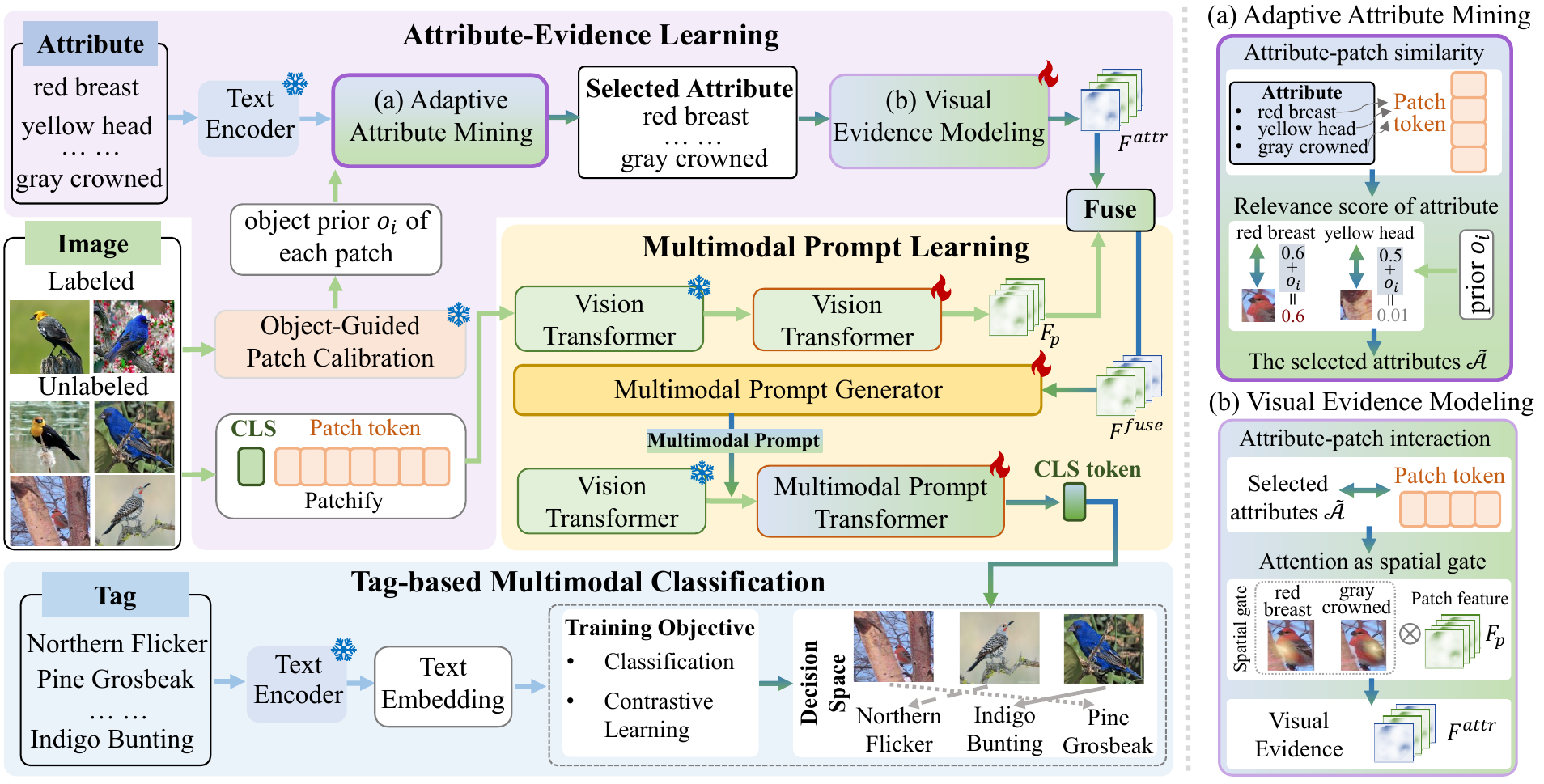}
\caption{\textbf{Overall framework of AptM-GCD.} AptM-GCD uses Attribute-Evidence Learning (AEL) and Multimodal Prompt Learning (MPL) to harness fine-grained attributes as apt textual guidance for representation refinement. AEL selects image-specific attributes and extracts attribute-conditioned visual evidence from object-aware patch tokens. 
MPL converts this evidence into multimodal prompts and injects them into the visual transformer to integrate local attribute cues with global visual context. The refined representations are finally used for tag-based multimodal classification with category tags as global semantic prototypes. The right panels provide detailed illustrations of adaptive attribute selection and visual evidence modeling.}
\label{fig:method}
\vspace{-10pt}
\end{figure*}

In addition to contrastive learning, the model adopts a parametric classifier represented by a set of learnable prototypes 
$\mathcal{C}=\{\boldsymbol{c}_1,\ldots,\boldsymbol{c}_J\}$, where $J$ denotes the number of class prototypes, corresponding to the total number of old and novel categories to be discovered.
Following the mean-teacher \cite{meanteacher} and self-distillation paradigm \cite{dino1,selfdis1}, the two augmented views are processed by the student and teacher networks, respectively. For an input view $\boldsymbol{x}_i$, its hidden feature $\boldsymbol{h}_i=F(\boldsymbol{x}_i)$ is compared with each prototype using cosine similarity. The predicted class distribution is obtained by a temperature-scaled softmax:
\begin{equation}
\label{equ1}
\boldsymbol{p}_i^{(j)}
=
\frac{
\exp \left(
\frac{1}{\tau_s}
\left(
\frac{\boldsymbol{h}_i}{\|\boldsymbol{h}_i\|_2}
\right)^\top
\left(
\frac{\boldsymbol{c}_j}{\|\boldsymbol{c}_j\|_2}
\right)
\right)
}{
\sum_{j^{\prime}}
\exp \left(
\frac{1}{\tau_s}
\left(
\frac{\boldsymbol{h}_i}{\|\boldsymbol{h}_i\|_2}
\right)^\top
\left(
\frac{\boldsymbol{c}_{j^{\prime}}}{\|\boldsymbol{c}_{j^{\prime}}\|_2}
\right)
\right)
}
\end{equation}
The teacher prediction from the other view, denoted as $\boldsymbol{q}_i^{\prime}$, is used as a pseudo label for unlabeled data. The unsupervised classification loss is then formulated as
$
\mathcal{L}_{\mathrm{cls}}^{u}
=
\frac{1}{|B|}
\sum_{i\in B}
\ell(\boldsymbol{q}_i^{\prime},\boldsymbol{p}_i),
$
where $\ell(\boldsymbol{q},\boldsymbol{p})=-\sum_j \boldsymbol{q}^{(j)}\log \boldsymbol{p}^{(j)}$ denotes the cross-entropy loss. For labeled samples, the ground-truth label $\boldsymbol{y}_i$ is used instead:
$
\mathcal{L}_{\mathrm{cls}}^{s}
=
\frac{1}{|B^l|}
\sum_{i\in B^l}
\ell(\boldsymbol{y}_i,\boldsymbol{p}_i).
$
To avoid degenerate solutions and encourage balanced cluster assignments, the unsupervised classification objective further includes a mean-entropy maximization regularizer \cite{selfdis1}:
$
H(\overline{\boldsymbol{p}})
=
-\sum_k \overline{\boldsymbol{p}}^{(j)}
\log \overline{\boldsymbol{p}}^{(j)},
\quad
\overline{\boldsymbol{p}}
=
\frac{1}{2|B|}
\sum_{i\in B}
(\boldsymbol{p}_i+\boldsymbol{p}_i^{\prime}).
$
The classification loss is given by
$
\mathcal{L}_{\mathrm{cls}}
=
\lambda
\left(
\mathcal{L}_{\mathrm{cls}}^{u}
-
\varepsilon H(\overline{\boldsymbol{p}})
\right)
+
(1-\lambda)
\mathcal{L}_{\mathrm{cls}}^{s}.
$

Finally, the overall GCD training objective combines the representation learning loss and the parametric classification loss:
$
\mathcal{L}_{\mathrm{GCD}}
=
\mathcal{L}_{\mathrm{rep}}
+
\mathcal{L}_{\mathrm{cls}} .
$

\section{The proposed method}

\subsection{Overview of AptM-GCD}
AptM-GCD is designed to harness rich attribute semantics as apt textual guidance for multimodal GCD. It consists of two key components: \textbf{Attribute-Evidence Learning (AEL)} and \textbf{Multimodal Prompt Learning (MPL)}.
AEL first grounds attributes into local visual evidence through patch-grounded attribute reasoning, where patch-level visual tokens select reliable attributes and the selected attributes further guide localized discriminative evidence extraction. 
MPL then converts the localized attribute evidence into soft visual prompts and injects them into the visual transformer, enabling fine-grained textual cues to interact with global visual context and refine visual representations. 
The refined visual representations are finally fed into the standard GCD classification for discovering both old and novel categories.

\begin{figure}[t]
\centering
\includegraphics[width=\linewidth]{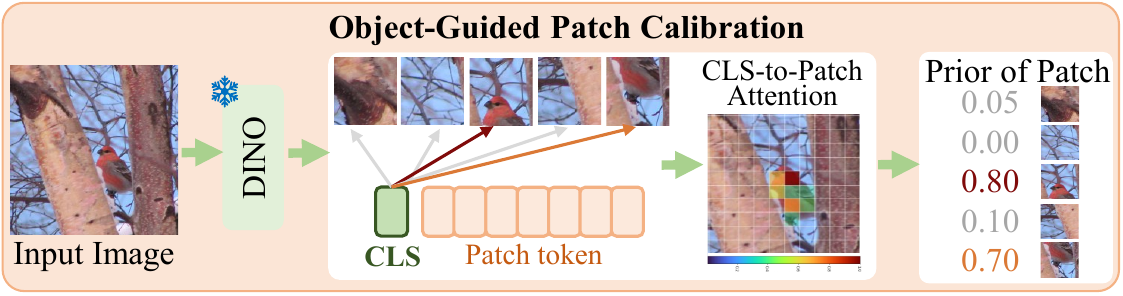}
\caption{\textbf{Object-Guided Patch Calibration.} We extract the last-layer \texttt{[CLS]}-to-patch attention from DINO and normalize it into a soft object prior. Each prior value measures the estimated object relevance of a patch, which is used to suppress background responses during adaptive attribute selection.}
\label{fig:method-obj}
\vspace{-10pt}
\end{figure}
\subsection{Attribute-Evidence Learning}
The Attribute-Evidence Learning (AEL) mechanism transforms attribute texts into fine-grained visual evidence through \textit{object-aware patch calibration}, \textit{adaptive attribute mining}, and \textit{visual evidence modeling}. 
Patch calibration suppresses background patches with soft object priors, adaptive selection identifies reliable image-specific attributes from patch-level responses, and evidence modeling uses the selected attributes to localize discriminative regions and refine patch-level representations.
For clarity, we summarize the overall procedure of this mechanism in Algorithm~\ref{alg:attribute_learning}.

\subsubsection{\textbf{Object-Guided Patch Calibration}}
Current multimodal GCD methods suffer from a granularity mismatch between global image representations and fine-grained textual attributes, as shown in Fig. \ref{fig:method-obj}. 
A natural remedy is to compare attributes with fine-grained visual features, such as patch tokens, so that local textual cues can be associated with localized visual evidence. 
However, not all patch-level features are discriminative. Similar to textual attributes, local visual tokens can also be noisy: not every patch captures discriminative object evidence, and background regions may spuriously respond to certain attributes.

To mitigate this issue, we introduce an object prior that estimates the relevance of each patch to foreground objects. Following prior observations that self-supervised DINO attention maps naturally highlight salient foreground regions \cite{dino1,object2}, we extract the last-layer \texttt{[CLS]}-to-patch attention as a training-free object prior, providing a weak cue for whether a patch is object-relevant.

Specifically, given an input image $x$, we first extract the last-layer self-attention tensor from the DINO encoder:
\begin{equation}
\mathbf{A} = \mathrm{DINO}_{\mathrm{attn}}(x) 
\in \mathbb{R}^{(N+1) \times (N+1)},
\end{equation}
where $N$ is the number of image patches. The additional token corresponds to the \texttt{[CLS]} token. We then obtain the \texttt{[CLS]}-to-patch attention scores $\mathbf{s} = \mathbf{A}_{0,1:N+1} \in \mathbb{R}^{N}$.

To construct a normalized soft object prior, we apply min-max normalization \cite{minmax}:
\begin{equation}
\hat{\mathbf{s}} =
\frac{
\mathbf{s} - \min_i s_{i}
}{
\max_i s_{i} - \min_j s_{j} + \epsilon
},
\end{equation}
where $\epsilon$ is a small constant for numerical stability. We further clamp the normalized scores and obtain the final soft object prior $\mathbf{o}\in\mathbb{R}^{N}$. 
Each element $o_i$ indicates the estimated object relevance of the $i^{th}$ patch and is later used as a soft weighting signal for adaptive attribute mining.

\subsubsection{\textbf{Adaptive Attribute Mining}}
To mine image-specific attributes more reliably, we perform patch-level retrieval instead of the global image-level retrieval commonly used in recent multimodal GCD methods. 
Furthermore, we incorporate the patch-level object prior obtained earlier to calibrate the similarity responses, preventing object-irrelevant background patches from dominating the attribute selection.

Let $\mathbf{P}=\{\mathbf{p}_{i}\}_{i=1}^{N}$ denote the patch features of CLIP, and $\mathcal{A}=\{\mathbf{a}_{m}\}_{m=1}^{M}$ the candidate attribute embeddings. 
We first compute the patch-attribute similarity matrix:
\begin{equation}
\mathbf{S}^{pa}(i,m) = \mathrm{sim}(\mathbf{p}_{i}, \mathbf{a}_{m}),
\end{equation}
where $\mathbf{S}^{pa} \in \mathbb{R}^{N \times M}$ contains the similarity scores between $N$ patches and $M$ attributes.

Directly using $\mathbf{S}^{pa}$ for attribute mining may still be affected by background patches, as each attribute could be spuriously activated by irrelevant regions. 
To mitigate this, we inject the object prior $\mathbf{o} \in \mathbb{R}^{N}$ as a log-prior bias into the patch-attribute similarity scores:

\begin{equation}
\tilde{\mathbf{S}}^{pa}(i,m)
=
\mathbf{S}^{pa}(i,m)
+
\log (o_{i}),
\end{equation}
The same prior bias is added to all attributes for a given patch. 
Consequently, object-relevant patches with $o_{i} \approx 1$ introduce almost no penalty, since $\log(o_{i}) \approx 0$, while background patches with small prior values receive a negative bias and are suppressed across all attributes.
We then aggregate the patch-level evidence for each attribute by taking the maximum response over all patches:

\begin{equation}
r_{m}
=
\max_{i \in \{1,\dots,N\}}
\tilde{\mathbf{S}}^{pa}(i,m),
\label{eq:attribute_score}
\end{equation}
where $r_m$ denotes the image-level relevance score of the $m$-th attribute.  
This operation evaluates each attribute by its single most supportive object-aware patch, using the corresponding maximum patch response as its relevance score to the current image.
The resulting scores provide the criterion for selecting image-specific attributes. 
To perform adaptive mining, we further normalize these relevance-scores into a probability distribution:
\begin{equation}
p_{m}
=
\frac{\exp(r_{m})}
{\sum_{m'=1}^{M} \exp(r_{m'})}.
\end{equation}
Let $\pi$ be a permutation that sorts the attributes in descending order of their probabilities:
\begin{equation}
p_{\pi(1)} \geq p_{\pi(2)} \geq \cdots \geq p_{\pi(M)}.
\end{equation}
We then perform nucleus selection by choosing the smallest attribute subset whose cumulative probability exceeds a threshold
\(
\rho\in(0,1]
\):
\begin{equation}
K
=
\min\left\{
k:
\sum_{m=1}^{k} p_{\pi(m)} \geq \rho
\right\}.
\label{equ:nucleus}
\end{equation}
The selected attribute index set is defined as:
\begin{equation}
\mathcal{I}
=
\left\{
\pi(m) \mid m=1,\dots,K
\right\},
\end{equation}
and the corresponding image-specific attribute feature set is:
\begin{equation}
\tilde{\mathcal{A}}
=
\left\{
\mathbf{a}_{f} \mid f \in \mathcal{I}
\right\}.
\end{equation}
In this way, attribute mining is driven by object-aware patch responses rather than global image-level similarity, reducing the influence of background-induced spurious correlations. The selected attribute set $\tilde{\mathcal{A}}$ therefore provides image-specific and object-relevant textual evidence for subsequent visual evidence modeling.

\begin{algorithm}[t]
\caption{Attribute-Evidence Learning Mechanism}
\label{alg:attribute_learning}
\begin{algorithmic}[1]
\STATE \textbf{Parameters:} attribute vocabulary $\mathcal{A}$, patch tokens $P$, patch feature map $F$, DINO attention $A$, nucleus threshold $\rho$, modulation $\lambda$
\STATE \textbf{Input:} image $x$
\STATE \textbf{Output:} attribute-guided feature map $F^{attr}$, selected attributes $\widetilde{\mathcal{A}}$

\STATE \textbf{Patch Calibration:}
\STATE $o \leftarrow \mathrm{ObjectPrior}(A)$ \COMMENT{patch relevance to foreground}

\STATE \textbf{Attribute Mining:}
\STATE $S^{pa} \leftarrow \mathrm{PatchAttributeSim}(P, \mathcal{A}) + \log(o)$
\STATE $r \leftarrow \mathrm{MaxPooling}(S^{pa})$
\STATE $\widetilde{\mathcal{A}} \leftarrow \mathrm{SelectTopAttributes}(r, \rho)$

\STATE \textbf{Evidence Modeling:}
\FOR{$a_{f} \in \widetilde{\mathcal{A}}$}
    \STATE $w_{f} \leftarrow \mathrm{Softmax}(r_{f})$ \COMMENT{attribute importance}
    \STATE $\mathrm{Cross}(r,:) \leftarrow \mathrm{Softmax}\Big((Q a_{f})^\top (K P) / \sqrt{d}\Big) \, V P$
\ENDFOR
\STATE $g \leftarrow \sum_{f} w_{f} \, \mathrm{Cross}(f,:)$ \COMMENT{aggregate cross-attention over attributes}
\STATE $G \leftarrow \mathrm{Reshape}(g)$
\STATE $F^{attr} \leftarrow F_p \odot (1+\lambda G)$

\RETURN $F^{attr}, \widetilde{\mathcal{A}}$
\end{algorithmic}
\end{algorithm}
% \vspace{-10pt}

\subsubsection{\textbf{Visual Evidence Modeling}}
The selected attributes provide image-specific semantic cues for the current image. 
We further use cross-attention~\cite{crossattn1,crossattn2} to retrieve their corresponding fine-grained visual evidence from patch tokens, treating attributes as semantic queries and patches as keys and values. 
This attribute-to-patch interaction retrieves localized evidence associated with reliable attributes, producing an evidence map that guides the model to focus on discriminative regional features.

Let
\(
\mathbf{P}=\{\mathbf{p}_{i}\}_{i=1}^{N}
\)
denote the patch tokens, \(
\tilde{\mathcal{A}} = \{\mathbf{a}_{f}\}_{f=1}^{K}
\)
indicate the selected attribute embeddings. 
The attribute-to-patch attention map is computed as:
\begin{equation}
\mathbf{Cross}(f,i)
=
\frac{
\exp \left(
\frac{
(\mathbf{W}_q \mathbf{a}_{f})^\top
(\mathbf{W}_k \mathbf{p}_{i})
}{
\sqrt{d}
}
\right)
}{
\sum_{j=1}^{N}
\exp \left(
\frac{
(\mathbf{W}_q \mathbf{a}_{f})^\top
(\mathbf{W}_k \mathbf{p}_{j})
}{
\sqrt{d}
}
\right)
},
\end{equation}
where
\(
\mathbf{Cross} \in \mathbb{R}^{K \times N}
\)
and each row
\(
\mathbf{Cross}(f,:)
\)
represents the spatial attention distribution of the $f^{th}$ selected attribute over all patches. 

Since different attributes contribute unequally to the visual representation, we weight each attribute's attention map by its patch-level similarity score, aggregating them into a unified patch relevance map that emphasizes the most informative attributes.
\begin{equation}
g_{i}
=
\sum_{f=1}^{K}
w_{f}\,\mathbf{Cross}(f,i), 
\end{equation}
where \begin{equation} w_{f} = \frac{\exp(r_{f})}{\sum_{f'=1}^{K} \exp(r_{f'})},\end{equation}
where $r_f$ is the relevance score of the selected attribute $f$, obtained from Eq.~(\ref{eq:attribute_score}). 
The resulting vector
\(
\mathbf{g} = \{g_i\}_{i=1}^{N}
\)
encodes the attribute-guided importance of each patch. 

By using attributes as semantic queries to attend over visual patches, it establishes a dynamic correspondence between textual semantics and spatial visual regions, enabling deeper multimodal fusion beyond global image-text alignment \cite{crossattn3}. 

This attribute-conditioned interaction further guides the model to emphasize task-relevant local details via reshaping $\mathbf{g}$ into a spatial gate $\mathbf{G} = \mathrm{reshape}(\mathbf{g}) \in \mathbb{R}^{1 \times H_p \times W_p}$ that aligns with the feature map $
\mathbf{F_{p}}
=
\mathrm{reshape}(\mathbf{P})
\in \mathbb{R}^{C \times H_p \times W_p}
$ reshaped by patch tokens.

We then modulate the visual feature map with the spatial gate in a residual manner, and
% \begin{equation}
% \mathbf{F}^{\mathrm{attr}}
% =
% \mathbf{F}
% \odot
% \left(
% 1 + \lambda \mathbf{G}
% \right),
% \end{equation}
% where $\odot$ denotes element-wise multiplication with broadcasting over the channel dimension.
% The scalar $\lambda$ controls the strength of attribute-guided modulation.
% In implementation, it corresponds to the product of two scaling factors:
% \begin{equation}
% \lambda = \alpha_{\mathrm{cp}} \cdot s_{\mathrm{cp}}.
% \end{equation}
the resulting feature map
\(
\mathbf{F}^{\mathrm{attr}}
\)
can be viewed as an attribute-aware recalibration of the original patch-level representation. 
In this way, attributes are no longer confined to the textual space but are integrated into the visual representation. This mechanism directs the model to focus on semantically salient regions while suppressing irrelevant background responses. 
Consequently, the visual evidence exhibit enhanced discriminability, enabling more precise differentiation between novel and old categories.

\subsection{Multimodal Prompt Learning}
After Attribute-Evidence Learning, the selected attributes are transformed into attribute-conditioned visual evidence, which captures fine-grained local cues supported by reliable textual semantics.
To incorporate such evidence into global visual reasoning, we introduce Multimodal Prompt Learning (MPL), a meta-learning-inspired prompting strategy \cite{aptgcd,metalearning} that learns to generate image-specific multimodal prompts rather than directly inserting attribute tokens or optimizing fixed prompts. 

MPL consists of two components: \textbf{Multimodal Prompt Generation}, which recurrently generates prompts from evidence, and \textbf{Multimodal Prompt Transformer}, which injects these prompts into the model to integrate local attribute evidence with global contextual features.
In this way, the model learns to prompt itself with reliable local evidence, enabling adaptive and context-aware representation learning for category discovery.

\begin{figure}[t]
\centering
\includegraphics[width=\linewidth]{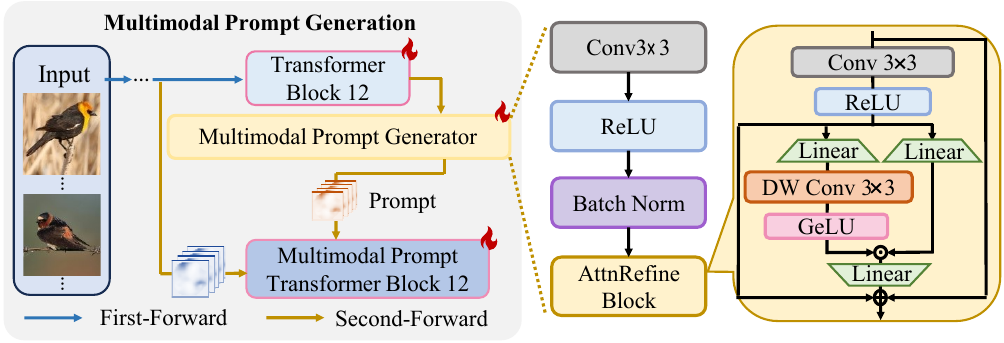}
\caption{\textbf{Multimodal Prompt Generation.} The left part shows the Recurrent Self-generation Mechanism, where the first forward pass constructs attribute-conditioned visual evidence through AEL and generates image-specific prompts that are reused in the second forward pass to guide the model. The right part shows the structure of the Multimodal Prompt Generator.}
\label{fig:method-prompt}
\vspace{-10pt}
\end{figure}

\subsubsection{\textbf{Multimodal Prompt Generation}}

Multimodal Prompt Generation consists of a \textit{Multimodal Prompt Generator} and a \textit{Recurrent Self-generation Mechanism}. 
The former generates multimodal prompts from the original patch-level visual feature map and the attribute-conditioned visual evidence, while the latter enables the prompts to be self-generated from the current image and reused to refine visual representation learning in a second forward pass.

Specifically, to construct prompts that encode both visual structure and attribute-supported local evidence, we use the original patch-level visual feature map
\(
\mathbf{F}_p \in \mathbb{R}^{C \times H_p \times W_p}
\)
and the attribute-conditioned visual evidence
\(
\mathbf{F}^{\mathrm{attr}} \in \mathbb{R}^{C \times H_p \times W_p}
\).
These two sources are first fused by channel-wise concatenation followed by a $1\times1$ convolution:
\begin{equation}
\mathbf{F}^{\mathrm{fuse}}
=
\phi_{\mathrm{fuse}}
\Bigl(
[\mathbf{F}_p ; \mathbf{F}^{\mathrm{attr}}]
\Bigr),
\end{equation}
where $\phi_{\mathrm{fuse}}$ is a learnable $1\times1$ convolution and $[\cdot;\cdot]$ denotes concatenation along the channel dimension.

The fused representation is then processed by a residual prompt generator inspired by AttnRefine~\cite{aptgcd}, as illustrated in Fig.~\ref{fig:method-prompt}. 
The generator incorporates a Gated Linear Unit (GLU)~\cite{glu} and depthwise convolution~\cite{deepwise} to selectively regulate channel-wise information flow and enhance local spatial details. 
Formally, the generated multimodal prompt is
\begin{equation}
\mathbf{MP}
=
\phi_{\mathrm{prompt}}(\mathbf{F}^{\mathrm{fuse}}),
\label{eq:MLP}
\end{equation}
where $\mathbf{MP} \in \mathbb{R}^{C \times H_p \times W_p}$ encodes both structural cues from the image and localized semantic evidence from selected attributes. 
The GLU adaptively controls channel-wise information propagation, while the depthwise convolution acts as a conditional positional encoding, allowing the prompt to capture nonlinear interactions and fine-grained spatial patterns.

The Recurrent Self-generation Mechanism further defines how the prompt is produced and used across two forward passes. 
In the first forward pass, the visual encoder produces intermediate patch representations, from which AEL selects object-aware attributes and constructs the attribute-conditioned visual evidence
\(
\mathbf{F}^{\mathrm{attr}}
\).
Based on the fused multimodal representation, the prompt generator produces
\(
\mathbf{MP}
\),
which is then reused in the second forward pass to guide the visual transformer. 
In this way, the prompt is not externally imposed or fixed, but self-generated from the current image and its reliable attribute evidence, enabling image-specific and evidence-aware guidance for representation learning.

\subsubsection{\textbf{Multimodal Prompt Transformer}}
The generated prompt $\mathbf{MP}$ in Eq.~\ref{eq:MLP} is injected into the second-pass visual transformer as additional queries in the final layers, enabling attribute-guided attention modulation alongside standard self-attention. 
Attribute-guided attention is performed via standard Query-Key-Value operations. $\mathbf{MP}$ is embedded as learnable-Q, added as a
prefix matrix \cite{prefix} to all query tokens, achieving similar
aggregation with minimal overhead.
This parallel fusion preserves the global perception of QKV attention while enhancing local feature modeling. 
The Final attention of the $i^{th}$ layer is formulated:
\begin{equation}
\boldsymbol{x_i^{att}}=\operatorname{softmax}\left(\frac{\left[Q_{x}, MP_q\right] K_{x}^T}{\sqrt{d}}\right) V_{x}.
\end{equation}
The output token $\boldsymbol{x_i^{att}}$ undergoes further processing through the LayerNorm and MLP layers, resulting in the final output $\boldsymbol{x_i} = MLP\left(LayerNorm\left(\boldsymbol{x_i^{att}}\right)\right) + \boldsymbol{x_i^{att}}$. 
Through this parallel prompt-attention design, the transformer preserves the global contextual modeling ability of standard QKV attention while biasing the refinement process toward object-specific and attribute-supported regions. 

By injecting attribute information into the visual transformer, the model can emphasize subtle object-specific evidence while still preserving the transformer’s ability to model global context \cite{TransZero++}.

Consequently, the final visual representation is enriched with localized semantic evidence mined by the attribute learning, which provides the fine-grained cues beyond coarse global image-level alignment. 
This is particularly beneficial for GCD, where novel categories may differ from known ones only by subtle attribute-level variations.

\subsection{Tag-based Multimodal Classification}
After representation refinement by our AEL and MPL, AptM-GCD follows the standard multimodal classification paradigm for category prediction \cite{spectralgcd,get,textgcd,ssr2gcd}. 
We use category tags as global semantic prototypes and classify images according to their similarities with tag prototypes.
Specifically, after the second forward pass guided by the multimodal prompt transformer, the visual encoder produces a refined global image representation
\(
\mathbf{z} \in \mathbb{R}^{D}
\)
for the input image.
Unlike the first-pass representation, this feature has already been refined by attribute-aware local evidence, and therefore contains both holistic visual context and fine-grained semantic cues.

Let
\(
\mathcal{T}=\{\mathbf{t}_m\}_{m=1}^{M_t}
\)
denote the global tag prototypes, where
\(
\mathbf{t}_m \in \mathbb{R}^{D}
\).
We normalize the refined image representation and the tag prototypes, and compute their cosine similarities:
$
\mathbf{c}_{m}
=
\frac{
\mathbf{z}^\top \mathbf{t}_m
}{
\|\mathbf{z}\|\;\|\mathbf{t}_m\|
},
$
where $c_{m}$ measures the semantic compatibility between the input image and the $m^{th}$ tag prototype. 

The resulting tag-image similarity vector provides the category-level prediction space for GCD. 
Based on this representation, the model is optimized with the contrastive and parametric classification losses described in Sec.~\ref{sec:pre}, enabling discovery of both old and novel categories.

\section{Experiments}
\label{sec:exper}

\begin{table}[t]
\centering
% 在 resizebox 内部嵌套一个 \mbox
\resizebox{\columnwidth}{!}{%
\mbox{%
\begin{threeparttable}
\caption{Dataset Specifications}\label{tab:dataset}
\setlength{\tabcolsep}{8pt}
\begin{tabular}{lccccc}
\toprule
& & \multicolumn{2}{c}{$\mathcal{D}_l$}      & \multicolumn{2}{c}{$\mathcal{D}_u$}  \\ \cmidrule(lr){3-4} \cmidrule(l){5-6} 
Dataset   & Balance    & \#Class & \#Img & \#Class           & \#Img \\ \midrule
CIFAR-10   & \ding{51}   & 5 & 12.5K  & 10          & 37.5K  \\
CIFAR-100     & \ding{51} & 80 & 20.0K & 100           & 30.0K \\
ImageNet-100 & \ding{51}   & 50 & 31.9K                                                               & 100            & 95.3K                                     \\
CUB       & \ding{51}   & 100  & 1.5K                                                              & 200               & 4.5K                                     \\
Stanford Cars & \ding{51} & 98 & 2.0K                                                       & 196 & 6.1K                                      \\
FGVC-Aircraft & \ding{51} & 50  & 1.7K & 100         & 5.0K \\
\bottomrule
\end{tabular}
\begin{tablenotes}[flushleft]
    \footnotesize \item[] Detailed information on the number of classes (\#Class) and images (\#Img) in labeled $\mathcal{D}_l$ and unlabeled $\mathcal{D}_u$. \textit{Balance} indicates whether the number of images per category in the dataset is balanced.
\end{tablenotes}
\end{threeparttable}%
}} % <--- 在这里依次闭合 \mbox 和 \resizebox
\end{table}

\subsection{Experimental Setup}
\begin{table}[t]
\caption{Comparison Results on fine-grained datasets}
\label{tab:ssb}
\centering
\begin{threeparttable}
\setlength{\tabcolsep}{2pt} 
\newcolumntype{Y}{>{\centering\arraybackslash}X}

% 方案二的关键：仅缩放 tabularx 主体，并用 \mbox 彻底解决 \prevdepth 报错
\resizebox{\columnwidth}{!}{%
\mbox{%
\begin{tabularx}{\columnwidth}{l YYY YYY YYY}
\toprule
\multirow{2}{*}{\textbf{Method}} 
& \multicolumn{3}{c}{\textbf{CUB}} 
& \multicolumn{3}{c}{\textbf{Stanford Cars}} 
& \multicolumn{3}{c}{\textbf{Aircraft}} \\
\cmidrule(lr){2-4} \cmidrule(lr){5-7} \cmidrule(lr){8-10}
& All & Old & New & All & Old & New & All & Old & New \\
\midrule
ClipGCD \cite{clipgcd} 
& 62.8 & 77.1 & 55.7 
& 70.6 & 88.2 & 62.2 
& 50.0 & 56.6 & 46.5 \\

TextGCD \cite{textgcd} 
& 76.6 & \textbf{80.6} & 74.7 
& 86.9 & 87.4 & 86.7 
& 50.8 & 44.9 & 53.8 \\

GET \cite{get}    
& 77.0 & 78.1 & 76.4 
& 78.5 & 86.8 & 74.5 
& 58.9 & 59.6 & 58.5 \\

SpectralGCD \cite{spectralgcd}    
& \underline{79.2} & \underline{80.4} & \underline{78.5} 
& 89.1 & 92.6 & \underline{87.4} 
& \underline{63.0} & \textbf{66.1} & \underline{61.4} \\

SSR$^2$-GCD \cite{ssr2gcd}    
& 78.3 & 78.5 & 78.2 
& \underline{89.2} & \underline{93.1} & 87.3 
& - & - & - \\
\midrule
\rowcolor[HTML]{E6ECF5} 
\textbf{AptM-GCD (Ours)} 
& \textbf{81.5} & \textbf{80.6} & \textbf{81.9} 
& \textbf{91.0} & \textbf{94.0} & \textbf{89.2} 
& \textbf{65.3} & \underline{65.5} & \textbf{65.3} \\
\bottomrule
\end{tabularx}%
}} % <--- 在这里依次闭合 \tabularx 和 \mbox、\resizebox

% 脚注在缩放盒子外面，字号不受影响，自动随单栏宽度换行
\begin{tablenotes}[flushleft]
\footnotesize
\item[] \textbf{Compared baseline:} ClipGCD (Arxiv'23), TextGCD (ECCV'24), GET (CVPR'25), SpectralGCD (ICLR'26), SSR$^2$-GCD (CVPR'26). \textbf{Bold values} indicate the best results, while \underline{underlined values} denote the second-best.
\end{tablenotes}
\end{threeparttable}
\vspace{-10pt}
\end{table}
\begin{table}[t]
\caption{Comparison Results on generic datasets}
\label{table:generic}
\centering
\begin{threeparttable}
\setlength{\tabcolsep}{2pt}
% 定义一个自动计算宽度且居中的 Y 列
\newcolumntype{Y}{>{\centering\arraybackslash}X}

% 关键改动：将 tabular 替换为 tabularx，并指定总宽为 \columnwidth
\resizebox{\columnwidth}{!}{%
\mbox{%
\begin{tabularx}{\columnwidth}{l YYY YYY YYY}
\toprule
\multirow{2}{*}{\textbf{Method}} 
& \multicolumn{3}{c}{\textbf{ImageNet-100}} 
& \multicolumn{3}{c}{\textbf{CIFAR-10}} 
& \multicolumn{3}{c}{\textbf{CIFAR-100}} \\
\cmidrule(lr){2-4} \cmidrule(lr){5-7} \cmidrule(lr){8-10}
& All & Old & New & All & Old & New & All & Old & New \\
\midrule
ClipGCD \cite{clipgcd}       
& 84.0 & 95.5 & 78.2 
& 96.6 & 97.2 & 96.4 
& 85.2 & 85.0 & 85.6 \\

TextGCD \cite{textgcd}        
& 91.7 & 95.7 & 89.7 
& 97.2 & 94.6 & 98.5 
& 82.1 & 85.5 & 75.5 \\

GET \cite{get}            
& 88.0 & 92.4 & 85.2 
& \underline{98.2} & \underline{98.0} & \underline{98.6} 
& 85.7 & 86.3 & 84.6 \\

SpectralGCD \cite{spectralgcd}  
& \underline{93.4} & \underline{96.1} & \underline{92.0} 
& \textbf{98.5} & 96.7 & \textbf{99.4} 
& 86.1 & \underline{87.2} & 83.9 \\

SSR$^2$-GCD \cite{ssr2gcd}     
& 92.1 & 96.0 & 90.2 
& \textbf{98.5} & \textbf{98.3} & \underline{98.6} 
& \underline{86.4} & 86.2 & \textbf{86.9} \\
\midrule
\rowcolor[HTML]{E6ECF5} 
\textbf{AptM-GCD (Ours)}                    
& \textbf{95.1} & \textbf{96.8} & \textbf{94.4} 
& \textbf{98.5} & 96.9 & \textbf{99.4} 
& \textbf{86.9} & \textbf{87.5} & \underline{85.7} \\
\bottomrule
\end{tabularx}%
}} % <--- 完美结束缩放盒子

% 此时 threeparttable 能够完美识别 \columnwidth 宽度，脚注对齐且自动换行
\begin{tablenotes}[flushleft]
\footnotesize
\item[] \textbf{Bold values} indicate the best results, while \underline{underlined values} denote the second-best.
\end{tablenotes}
\end{threeparttable}
\vspace{-10pt}
\end{table}

\textbf{Dataset.} 
We evaluate AptM-GCD on six benchmark datasets, covering both fine-grained and general image recognition scenarios. The fine-grained datasets, from the Semantic Shift Benchmark (SSB) \cite{ssb}, include CUB \cite{cub}, Stanford Cars \cite{cars}, and FGVC-Aircraft \cite{aircraft}. The general recognition datasets comprise CIFAR-10/100 \cite{cifar} and ImageNet-100 \cite{imagenet}. Detailed dataset splits are summarized in Tab. \ref{tab:dataset}.

\textbf{Evaluation protocol.} 
Following common practice \cite{gcd}, we use clustering accuracy (ACC) to measure the quality of the predicted assignments. Given the predicted cluster label $y_i$ and the ground-truth label $\widehat{y}_i$, ACC is computed as
$\mathrm{ACC}=\frac{1}{|\mathcal{D}_u|}\sum_{i=1}^{|\mathcal{D}_u|}
\mathds{1}\left(y_i=\mathcal{P}(\widehat{y}_i)\right),$
where $\mathcal{P}$ denotes the optimal one-to-one mapping obtained by the Hungarian algorithm \cite{hungarian}. We report ACC separately for Old categories, New categories, and their union (All).

\textbf{Implementation details.}
Following prior multimodal GCD methods \cite{textgcd, spectralgcd, ssr2gcd}, we adopt CLIP ViT-B/16 \cite{clip} as the visual encoder and fine-tune only its last transformer block. The text encoder is kept frozen and used solely to extract embeddings from the tag and attribute dictionary provided by TextGCD \cite{textgcd}.
Models are trained for 200 epochs with a batch size of 128 \cite{spectralgcd}. Cosine learning rate annealing is applied with separate initial rates: $5\times10^{-3}$ for the visual backbone and $0.1$ for the remaining learnable modules, including the classifier, MLP, and linear projection layers. All experiments are conducted on a single NVIDIA RTX 3090 GPU.

\subsection{Comparison with State-of-the-Art Methods}

We compare AptM-GCD with SOTA multimodal GCD methods on both fine-grained and generic benchmarks. 
\subsubsection{\textbf{Evaluation on fine-grained datasets}}
As shown in Tab.~\ref{tab:ssb}, across all three fine-grained datasets, AptM-GCD achieves the best overall performance, demonstrating strong capability in modeling subtle inter-class variations. The improvement is particularly significant on novel categories, where AptM-GCD consistently obtains the highest New accuracy.
On FGVC-Aircraft, AptM-GCD improves New accuracy by 3.9\% over the strongest baseline, highlighting the effectiveness of attribute-conditioned visual evidence for fine-grained category discovery. Although the Old accuracy is slightly lower than SpectralGCD, AptM-GCD achieves the best All accuracy, indicating a better trade-off between preserving knowledge of seen classes and discovering novel categories.

\subsubsection{\textbf{Evaluation on generic datasets}}
Tab.~\ref{table:generic} further reports experimental results on generic benchmarks. AptM-GCD achieves the best All accuracy on both ImageNet-100 and CIFAR-100. In particular, it improves New accuracy to 94.4\% on ImageNet-100, surpassing the strongest baseline by 2.4\%, and reaches 86.9\% All accuracy on CIFAR-100.
These results demonstrate that AptM-GCD is not limited to fine-grained scenarios but generalizes effectively to generic tasks. The consistent improvements across all six datasets validate the benefit of modeling attributes as localized visual evidence and integrating them through multimodal prompt learning.

\section{Ablation Studies and Analysis}

\subsection{Effect of the AptM-GCD Framework}
We evaluate how AptM-GCD leverages attributes as reliable fine-grained guidance from two complementary perspectives. 
For attribute-based multimodal GCD methods, we assess whether AptM-GCD mitigates spurious associations caused by aligning local attributes with global image representations, which can bias predictions toward incorrect classes. 
For tag-based methods, we examine whether grounding attributes as localized visual evidence enables effective use of rich attribute semantics, rather than treating them as additional global textual cues. 
These studies demonstrate that the value of attributes lies not in simply adding textual information, but in using it appropriately to guide category discovery.

To ensure a fair comparison of textual resources, all attribute-related experiments in this section use the same tag and attribute dictionary provided by TextGCD \cite{textgcd}. Specifically, we conduct controlled comparisons on both the attribute-based TextGCD and the tag-based SpectralGCD \cite{spectralgcd}, keeping the available textual vocabulary identical while varying only how attribute information is exploited.

\begin{table}[t]
\centering
\begin{threeparttable}
\caption{Quantitative analysis of AptM-GCD on attribute-based method \cite{textgcd}}
\label{tab:ab-branch}
\setlength{\tabcolsep}{1.2pt} 
\setlength{\extrarowheight}{1pt} 

\begin{tabular}{cccccccccc}
\toprule
\multicolumn{5}{c}{\cellcolor{cubhead}\textbf{CUB}} & \multicolumn{5}{c}{\cellcolor{airhead}\textbf{Aircraft}} \\
\cmidrule(r){1-5} \cmidrule(l){6-10}
\cellcolor{cubhead}All $\uparrow$ & 
\cellcolor{cubhead}Old $\uparrow$ & 
\cellcolor{cubhead}New $\uparrow$ & 
\cellcolor{cubhead}M $\downarrow$(MB) & 
\cellcolor{cubhead}T $\downarrow$(s) & 
\cellcolor{airhead}All $\uparrow$ & 
\cellcolor{airhead}Old $\uparrow$ & 
\cellcolor{airhead}New $\uparrow$ & 
\cellcolor{airhead}M $\downarrow$(MB) & 
\cellcolor{airhead}T $\downarrow$(s) \\
\midrule

\multicolumn{10}{c}{\textit{\textcolor{black!55}{TextGCD w. tag}}} \\ 

69.9 & 74.9 & 67.3 & \textbf{5256} & \textbf{48} & 48.1 & 46.3 & 49.0 & \textbf{5256} & \textbf{52} \\
\midrule

\multicolumn{10}{c}{\textit{\textcolor{black!55}{TextGCD w. tag + attribute}}}  \\ 

76.6 & 80.6 & 74.7 & 7038 & 62 & 50.8 & 44.9 & 53.8 & 7038 & 66 \\
\midrule

\multicolumn{10}{c}{\textit{\textcolor{black!55}{TextGCD w. our framework}}}  \\ 

\rowcolor[HTML]{E6ECF5} 
\textbf{78.4} & \textbf{80.4} & \textbf{77.3} & 8386 & 54 & \textbf{53.1} & \textbf{49.3} & \textbf{55.7} & 8364 & 56 \\
\bottomrule
\end{tabular}

\begin{tablenotes}[flushleft]
\footnotesize
\item[] \textbf{\textit{M}} denotes GPU memory usage in MB, and \textbf{\textit{T}} denotes the training time per epoch in seconds.
\end{tablenotes}
\end{threeparttable}
\end{table}
\begin{figure}[t]
\centering
\includegraphics[width=\linewidth]{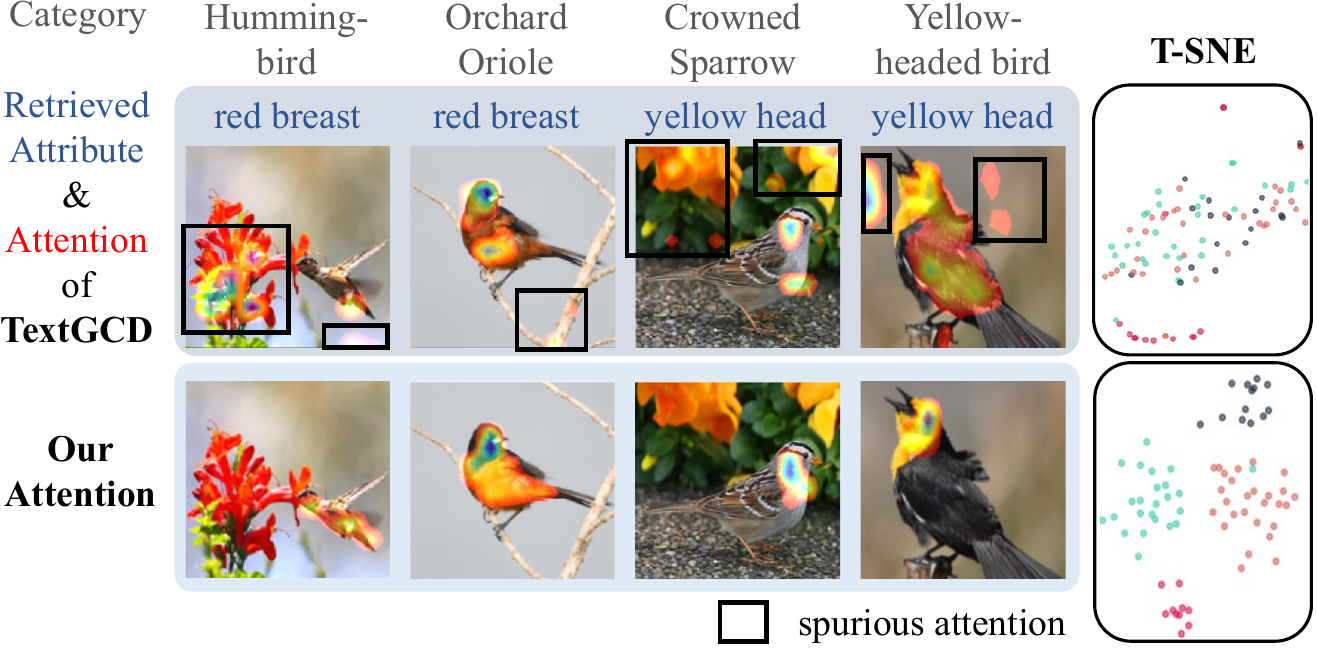}
\caption{\textbf{Qualitative analysis of AptM-GCD on attribute-based methods.} Last-layer attention maps and t-SNE feature distributions show that directly using retrieved attributes can cause spurious attention and entangled category boundaries. AptM-GCD alleviates this issue by transforming attributes into reliable visual evidence, guiding the model to focus on discriminative object regions and produce clearer category separation.}
\label{fig:ab-branch-fig}
\vspace{-10pt}
\end{figure}

% \begin{table}[t]
% \centering
% \caption{Backbone contribution in AptM-GCD}
% \label{tab:ab-component}
% \setlength{\tabcolsep}{4.5pt} % 适当微调列间距以完美适应 11 列排版
% \resizebox{\columnwidth}{!}{%
% \begin{tabular}{cccccccccc}
% \toprule
% \multirow{2}{*}{\textbf{Method}}  & \multicolumn{3}{c}{\textbf{CUB}} & \multicolumn{3}{c}{\textbf{Stanford Cars}} & \multicolumn{3}{c}{\textbf{Aircraft}} \\ 
% \cmidrule(r){2-4} \cmidrule(lr){5-7} \cmidrule(l){8-10}
%  & All & Old & New & All & Old & New & All & Old & New \\ 
% \midrule
% CLIP w/o. AEL\&MPL  & 78.0 & 80.5 & 76.4 & 87.0 & 93.3 & 84.2 & 61.3 & 65.4 & 59.3 \\ 
% \rowcolor[HTML]{E6ECF5} 
% CLIP w. AEL\&MPL &\textbf{81.5} & \textbf{80.6} & \textbf{81.9} & \textbf{91.0} & \textbf{94.0} & \textbf{89.2} & \textbf{65.3} & \textbf{65.5} & \textbf{65.3} \\ 
% \bottomrule
% \end{tabular}%
% }
% \end{table}

\subsubsection{\textbf{Attribute Usage in Attribute-Aware Methods}}
To evaluate how AptM-GCD improves the use of attributes in attribute-aware GCD methods, 
we integrate our AptM-GCD into TextGCD~\cite{textgcd}.
Fig.~\ref{fig:ab-branch-fig} visualizes retrieved attributes, last-layer attention maps, and t-SNE feature distributions. 
Directly using retrieved attributes together with tags as image-level textual descriptions can induce spurious semantic associations: samples from different categories may be grouped together when they share similar but non-discriminative attributes, and attention often activates on irrelevant or background regions. This confusion is further reflected in the t-SNE plots, where categories become entangled due to attribute-induced associations.
With AptM-GCD, retrieved attributes are transformed into localized visual evidence and injected into representation learning via multimodal prompts. 
This prevents misleading attributes from biasing category predictions and guides the model to focus on discriminative object regions. Quantitatively, our framework achieves the best results among all TextGCD variants in Tab.~\ref{tab:ab-branch}, with only moderate additional GPU memory (M) and training cost (T) per epoch.

\subsubsection{\textbf{Attribute Usage in Tag-Based Methods}}
For tag-based multimodal GCD methods such as SpectralGCD~\cite{spectralgcd}, we investigate whether attributes can be effectively incorporated as additional semantic cues. Specifically, we construct a naive attribute-enhanced baseline by augmenting the original tag vocabulary with the attribute vocabulary provided by TextGCD~\cite{textgcd}. Both tags and attributes are then matched against global image representations and used as textual guidance.
As shown in Fig.~\ref{fig:intro}(c), directly incorporating attributes into the tag vocabulary consistently degrades performance, suggesting that global image-level matching is insufficient for exploiting fine-grained attribute semantics. In contrast, our AptM-GCD achieves consistent improvements over the original tag-based baseline. By grounding attributes as localized visual evidence, AptM-GCD enables attribute information to complement category-level semantics without interfering with category discovery. These results demonstrate that the effectiveness of attributes depends on how they are incorporated, rather than simply increasing the amount of textual guidance.

\begin{table}[t]
\centering
\caption{Component ablation of AEL and MPL in AptM-GCD}
\label{tab:ab-component}
\setlength{\tabcolsep}{4.5pt} % 适当微调列间距以完美适应 11 列排版
\resizebox{\columnwidth}{!}{%
\begin{tabular}{ccccccccccc}
\toprule
\multirow{2}{*}{\textbf{AEL}} & \multirow{2}{*}{\textbf{MPL}} & \multicolumn{3}{c}{\textbf{CUB}} & \multicolumn{3}{c}{\textbf{Stanford Cars}} & \multicolumn{3}{c}{\textbf{Aircraft}} \\ 
\cmidrule(r){3-5} \cmidrule(lr){6-8} \cmidrule(l){9-11}
 &  & All & Old & New & All & Old & New & All & Old & New \\ 
\midrule
\ding{55} & \ding{55} & 78.0 & 80.5 & 76.4 & 87.0 & 93.3 & 84.2 & 61.3 & 65.4 & 59.3 \\ 
\ding{51} & \ding{55} & 79.2 & 80.4 & 78.6 & 88.8 & 93.5 & 86.5 & 63.3 & 65.9 & 62.0 \\ 
\ding{55} & \ding{51} & 80.1 & 81.1 & 79.6 & 90.2 & 93.6 & 88.4 & 64.7 & 65.7 & 64.3 \\ 

\rowcolor[HTML]{E6ECF5} 
\ding{51} & \ding{51} & \textbf{81.5} & \textbf{80.6} & \textbf{81.9} & \textbf{91.0} & \textbf{94.0} & \textbf{89.2} & \textbf{65.3} & \textbf{65.5} & \textbf{65.3} \\ 
\bottomrule
\end{tabular}%
}
\end{table}
\begin{figure}[t]
\centering
\includegraphics[width=\linewidth]{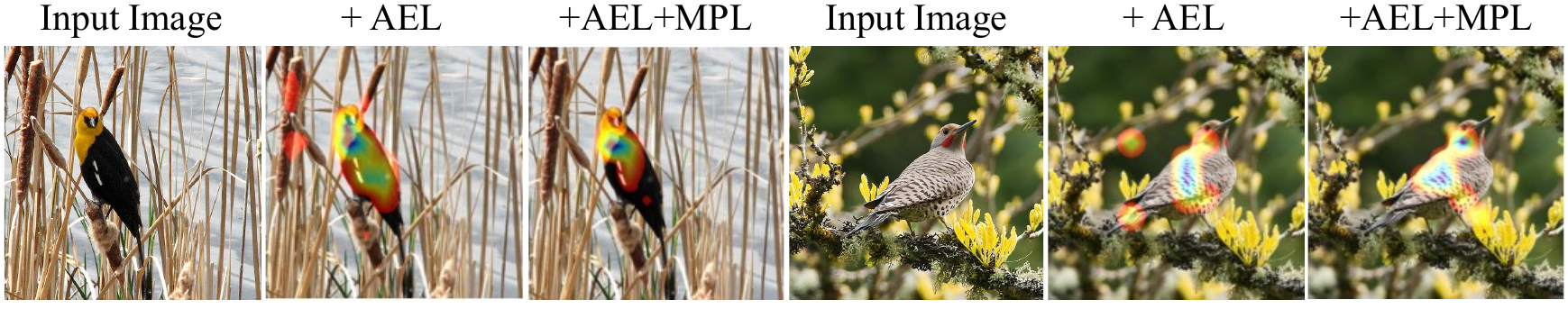}
\caption{\textbf{Qualitative analysis of component in AptM-GCD.} AEL localizes attribute evidence, while MPL integrates such local evidence with global visual context through soft visual prompts, leading to more reliable attention.}
\label{fig:ab-component-fig}
\vspace{-10pt}
\end{figure}

\subsection{Effect of the component in AptM-GCD}
We analyze the contribution of AEL and MPL in the overall framework by using the tag-based method~\cite{spectralgcd} with naively added attributes as the baseline. As reported in Tab.~\ref{tab:ab-component}, both components bring clear improvements over the baseline, and their combination achieves the best performance across all datasets. In particular, AEL and MPL together improve the novel-class accuracy by \textbf{5.5\%} on CUB, \textbf{5.0\%} on Stanford Cars, and \textbf{6.0\%} on FGVC-Aircraft. These results indicate that simply adding attributes to a tag-based GCD method is insufficient, while properly grounding and integrating attribute information can provide reliable fine-grained guidance for novel category discovery.

The qualitative results in Fig. \ref{fig:ab-component-fig} further explain the role of each component. AEL transforms mined attributes into localized visual evidence and reduces spurious associations with non-object regions. For example, it can highlight fine-grained object cues such as a bird's yellow head. However, using AEL alone is still vulnerable to visually similar background patterns, because the localized evidence is not fully integrated with global visual context. MPL addresses this limitation by converting attribute-conditioned visual evidence into multimodal prompts and injecting them into the visual transformer. In this way, local attribute cues can interact with global context during representation learning, allowing the model to better interpret context-dependent visual signals and produce more discriminative representations.

\begin{table}[t]
\centering
\caption{Quantitative ablation of Object-guided Patch Calibration}
\label{tab:ab-prior}
\setlength{\tabcolsep}{6pt} % 稍微缩小列间距以适应 10 列的排版
\resizebox{\columnwidth}{!}{%
\begin{tabular}{cccccccccc}
\toprule
\multirow{2}{*}{\textbf{Prior}} & \multicolumn{3}{c}{CUB} & \multicolumn{3}{c}{Stanford Scars} & \multicolumn{3}{c}{Aircraft} \\ 
\cmidrule(r){2-4} \cmidrule(lr){5-7} \cmidrule(l){8-10}
 & All & Old & New & All & Old & New  & All & Old & New\\ 
\midrule
\ding{55} & 80.3 & 80.5 & 80.2 & 90.7 & 93.8 & 88.6 & 64.0 & 65.3 & 63.3 \\ 
\rowcolor[HTML]{E6ECF5} 
\ding{51} & \textbf{81.5} & \textbf{80.6} & \textbf{81.9} & \textbf{91.0} & \textbf{94.0} & \textbf{89.2} & \textbf{65.3} & \textbf{65.5} & \textbf{65.3} \\ 
\bottomrule
\end{tabular}%
}
\end{table}
\begin{figure}[t]
\centering
\includegraphics[width=\linewidth]{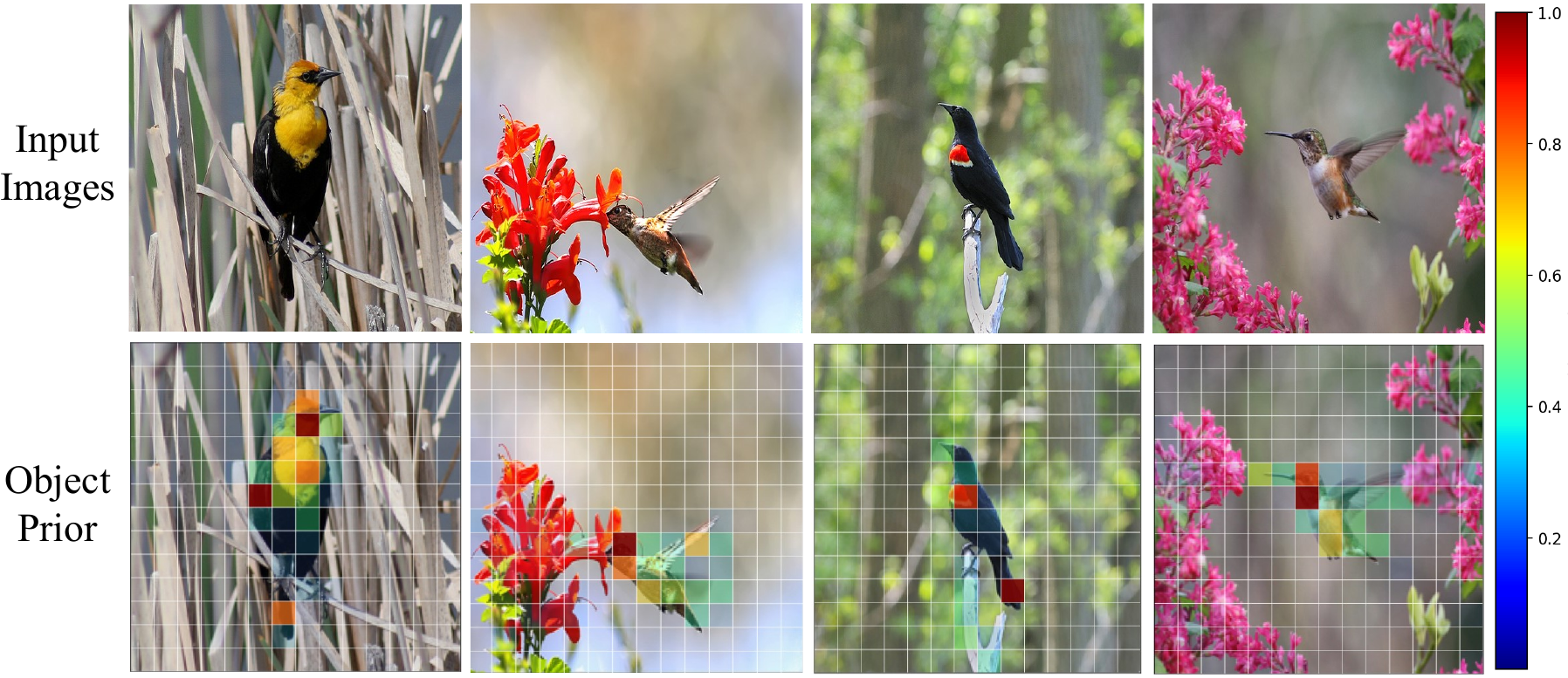}
\caption{\textbf{Visualization of the object prior.} Patch colors indicate the estimated object relevance, changing from red to blue as the prior value decreases. 
The prior highlights foreground object regions while suppressing background patches, enabling more reliable attribute selection.}
\label{fig:ab-prior-fig}
\vspace{-10pt}
\end{figure}

\begin{table}[t]
\centering
\caption{revisionred}{Robustness and cost comparison of different object prior sources.}
\label{tab:prior_source_cost}
\setlength{\tabcolsep}{1pt}
\renewcommand{\arraystretch}{1.12}
\resizebox{\columnwidth}{!}{%
\begin{tabular}{cccccccccccc}
\toprule
 \multicolumn{4}{c}{\cellcolor{cubhead}\textbf{CUB}}
& \multicolumn{4}{c}{\cellcolor{lightbg}\textbf{Stanford Scars}}
& \multicolumn{4}{c}{\cellcolor{airhead}\textbf{Aircraft}} \\
\cmidrule(lr){1-4}
\cmidrule(lr){5-8}
\cmidrule(lr){9-12}
\cellcolor{cubhead} All $\uparrow$ &\cellcolor{cubhead} New $\uparrow$ & \cellcolor{cubhead} T(s) $\downarrow$ & \cellcolor{cubhead} M(MB) $\downarrow$

&\cellcolor{lightbg} All $\uparrow$ & \cellcolor{lightbg} New $\uparrow$ & \cellcolor{lightbg} T(s) $\downarrow$ & \cellcolor{lightbg} M(MB) $\downarrow$

&\cellcolor{airhead} All $\uparrow$ & \cellcolor{airhead} New $\uparrow$ & \cellcolor{airhead} T(s) $\downarrow$ & \cellcolor{airhead} M(MB) $\downarrow$ \\
\midrule

\multicolumn{12}{c}{\textit{\textcolor{black!55}{CLIP self-prior}}} \\ 

80.4 & 80.8 & \textbf{0.05} & \textbf{148}
& 90.7 & 88.6 & \textbf{0.34} & \textbf{148}
& 64.4 & 63.9 & \textbf{0.52} & \textbf{148} \\

\multicolumn{12}{c}{\textit{\textcolor{black!55}{U2-Net saliency prior}}} \\

81.0 & 81.4 & 12 & 1260
& 90.8 & 88.8 & 23 & 1260
& 64.7 & 64.9 & 30 & 1260 \\

\multicolumn{12}{c}{\textit{\textcolor{black!55}{DINO attention prior}}} \\ 

\rowcolor[HTML]{E6ECF5} 
\textbf{81.5}  & \textbf{81.9}  & 19 & 1630
& \textbf{91.0}  & \textbf{89.2}  & 28 & 1630
& \textbf{65.3} & \textbf{65.3} & 37 & 1630 \\
\bottomrule
\end{tabular}%
}
\end{table}

\subsection{Ablation study on the Attribute-Evidence Learning}
To analyze the contribution of AEL in AptM-GCD, we conduct detailed ablations on its key components, including \textbf{Object-guided Patch Calibration}, \textbf{Adaptive Attribute Mining}, and \textbf{Visual Evidence Modeling}.

\subsubsection{\textbf{Effect of Object-Guided Patch Calibration}}
We evaluate this component through both quantitative ablation and qualitative visualization. As shown in Tab.~\ref{tab:ab-prior}, removing the object prior leads to a 2.8–5.1\% drop in ACC on novel classes across CUB, Cars, and Aircraft, highlighting the importance of reliable fine-grained cues for unlabeled novel categories. 
Fig.~\ref{fig:ab-prior-fig} visualizes the estimated object prior by overlaying patch grids on input images, where patch colors gradually change from red to blue as prior values decrease. High-prior patches focus on foreground objects, suppressing background regions even when they contain attribute-like cues, ensuring AEL mines discriminative local evidence. Salient object parts, such as discriminative regions in the first and third images, receive particularly strong responses.
These results demonstrate that object-guided patch calibration effectively filters out noisy background cues, allowing AEL to extract reliable fine-grained evidence for novel-category discovery.

We further investigate the robustness and cost of different object prior sources in Tab.~\ref{tab:prior_source_cost}. 
Here, CLIP self-prior denotes a lightweight internal prior computed from the similarity between the \texttt{[CLS]} token and patch tokens of the original CLIP visual encoder, U2-Net \cite{u2net} saliency prior uses an external salient object detection model to estimate foreground regions.
Among them, DINO achieves the best performance on all datasets, suggesting that DINO attention provides more reliable object-level localization for fine-grained attribute selection. CLIP self-prior has the lowest computational overhead because it reuses the visual tokens from the original CLIP encoder without introducing an additional model, but its performance is slightly lower since CLIP features are more category-semantic rather than explicitly foreground-localized. 
U2-Net saliency prior offers a practical trade-off, achieving competitive accuracy with lower memory cost than DINO. These results demonstrate that object-guided patch calibration is robust to different foreground prior sources, while DINO provides the accuracy-efficiency balance in our framework.

\begin{figure}[t]
\centering
\includegraphics[width=\linewidth]{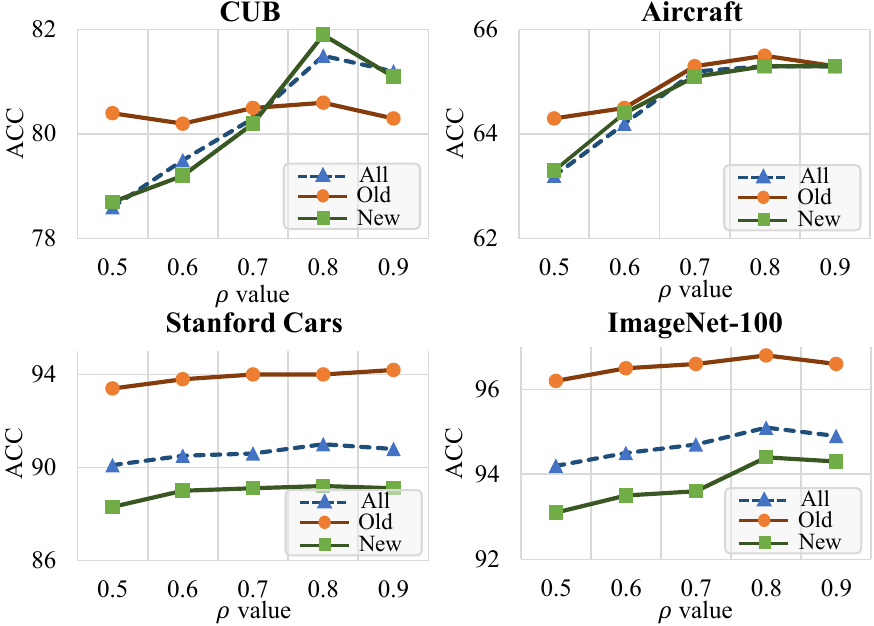}
\caption{\textbf{Sensitivity of the nucleus threshold $\rho$ in Adaptive Attribute Selection.} We vary $\rho$ from 0.5 to 0.9 on three fine-grained datasets and ImageNet-100. $\rho=0.8$ providing the best balance between attribute coverage and selection reliability}
\label{fig:ab-filter}
\end{figure}

\begin{table}[t]
\centering
\caption{Average number of selected attributes under different nucleus thresholds $\rho$.}
\label{tab:ab-numattr}
\setlength{\tabcolsep}{15pt} 
\begin{tabular}{c|ccc}
\hline
\(\rho\) & CUB & Stanford Cars & Aircraft \\
\hline
0.5 & 14 & 9  & 6  \\
0.6 & 16 & 11 & 8  \\
0.7 & 23 & 13 & 9  \\
\rowcolor[HTML]{E6ECF5} 
0.8 & 29 & 15 & 10 \\
0.9 & 37 & 17 & 11 \\
\hline
\end{tabular}
\end{table}

\begin{figure}[t]
\centering
\includegraphics[width=\linewidth]{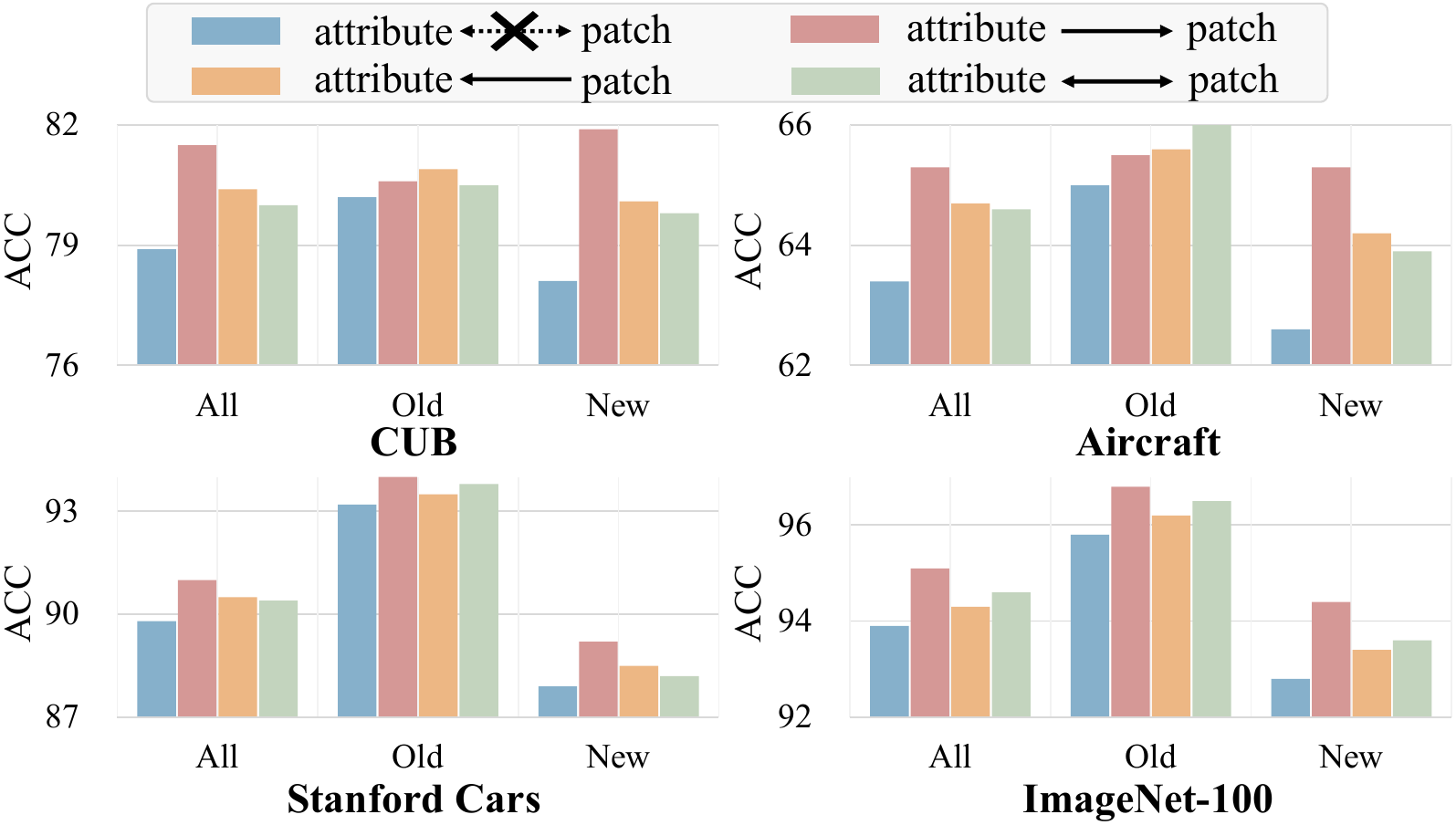}
\caption{\textbf{Effect of attribute-patch interaction strategies in Visual Evidence Modeling.} We compare different ways of modeling the correspondence between selected attributes and visual patches. Arrows denote the query direction in cross-attention, where attribute $\rightarrow$ patch uses attributes as queries to retrieve visual evidence from patches.}
\label{fig:ab-interaction}
\vspace{-10pt}
\end{figure}

\subsubsection{\textbf{Sensitivity to the Nucleus Threshold}}
Keeping all other components fixed, we evaluate the sensitivity of the nucleus threshold $\rho$ in Eq.~\ref{equ:nucleus} by varying it from 0.5 to 0.9 on three fine-grained datasets and ImageNet-100.
As shown in Fig.~\ref{fig:ab-filter}, AptM-GCD remains robust across a broad range of threshold values, indicating that Adaptive Attribute Mining is not overly sensitive to the choice of $\rho$. Performance generally improves as $\rho$ increases from 0.6 to 0.8, suggesting that retaining a moderately larger set of candidate attributes provides richer fine-grained evidence. However, further increasing $\rho$ to 0.9 yields marginal gains, as additional low-confidence attributes contribute limited discriminative information and may introduce redundancy. Based on these observations, we use $\rho=0.8$ in all experiments.

To further clarify how the nucleus threshold affects the actual number of selected attributes, we report the average number of retained attributes under different $\rho$ values in Tab. \ref{tab:ab-numattr}.
As $\rho$ increases from 0.5 to 0.9, the average number of selected attributes consistently increases across all datasets.
Together with Fig. \ref{fig:ab-filter}, this shows that selecting more attributes generally provides richer fine-grained evidence and improves category discovery performance up to a moderate range.
However, when $\rho$ becomes too large, additional low-confidence attributes may introduce redundant or noisy cues, leading to saturated or marginally degraded performance.

\subsubsection{\textbf{Interaction Strategy in Visual Evidence Modeling}}
To examine how mined attributes should interact with visual patches during evidence modeling, we compare four interaction strategies while keeping all other components fixed: no interaction, attribute-as-query, patch-as-query, and bidirectional interaction. In the no-interaction variant, selected attributes are simply averaged and passed to the prompt generator without patch-level interaction.
As shown in Fig.~\ref{fig:ab-interaction}, the attribute-as-query strategy achieves the best performance. This design treats attributes as semantic queries and patch tokens as candidate evidence, allowing the model to retrieve object-relevant visual regions that support the mined attributes. In contrast, the patch-as-query strategy requires each patch to search for relevant attributes, which is less aligned with the objective of evidence extraction and tends to resemble patch-level attribute classification. Although bidirectional interaction allows information exchange in both directions, it introduces additional complexity without improving performance. These results suggest that using attributes as semantic queries and visual patches as evidence providers is the most effective strategy for constructing reliable attribute-conditioned evidence.

\subsection{Ablation on Multimodal Prompt Learning}
We perform detailed ablations on MPL to evaluate its role within AptM-GCD, quantify the contribution of each prompt component, and assess its generalization as a vision prompt learning strategy for unimodal and class-imbalanced GCD without textual guidance.

% \subsubsection{\textbf{Effect of MPL in AptM-GCD}}
% We first evaluate the contribution of MPL within AptM-GCD through both qualitative and quantitative analyses (Fig.~\ref{fig:ab-prompt-all}). While AEL transforms mined attributes into localized visual evidence and reduces spurious associations with non-object regions—highlighting fine-grained cues such as a bird's yellow head—using AEL alone still leaves the model susceptible to visually similar background patterns. This is because local evidence is not fully integrated with global context, as illustrated in Fig.~\ref{fig:ab-prompt-all}(a).
% MPL addresses this limitation by converting attribute-conditioned visual evidence into multimodal prompts and injecting them into the visual transformer. This enables local attribute cues to interact with global visual context during representation learning, helping the model correctly interpret context-dependent visual signals.
% Quantitatively, using the tag-based method~\cite{spectralgcd} with naively added attributes as the baseline, Fig.~\ref{fig:ab-prompt-all}(b) shows that MPL substantially improves novel-class accuracy. This confirms that fine-grained attribute evidence is most effective when integrated with global context through prompt-guided representation learning.

\begin{figure}[t]
\centering
\includegraphics[width=\linewidth]{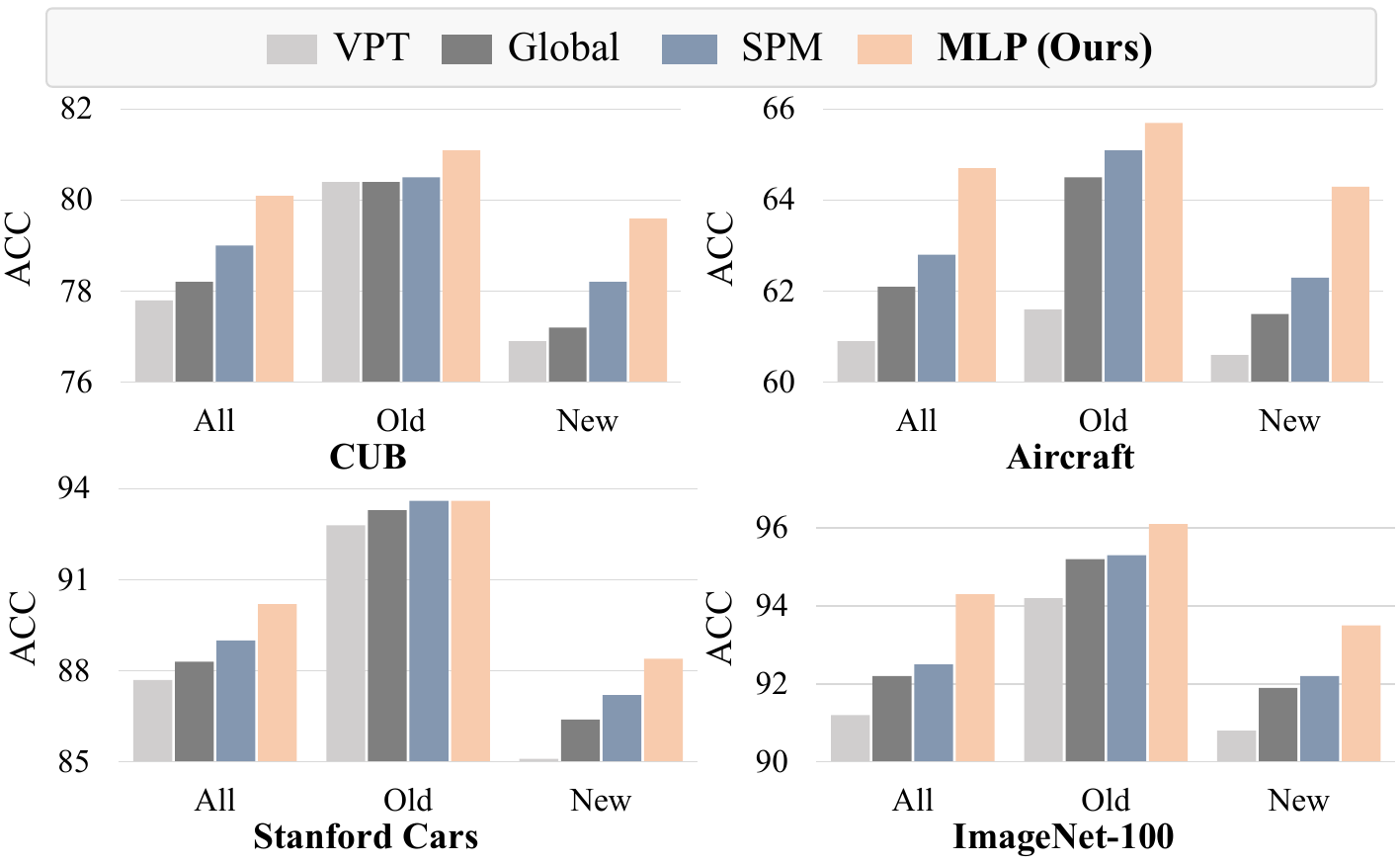}
\caption{\textbf{Effect of MPL as a visual prompting method within AptM-GCD.} Without AEL or textual attributes, MPL outperforms representative visual prompt methods, including VPT \cite{vpt}, Global \cite{global}, and SPM \cite{spm}, showing its strong ability to refine visual representations. Adding AEL further improves performance.}
\label{fig:ab-vp-our}
\vspace{-10pt}
\end{figure}

\subsubsection{\textbf{Effect of MPL as a Visual Prompt Method}}
\label{sec:exp-mpl}
We further investigate MPL as a standalone visual prompting strategy, independent of AEL or textual attributes, to isolate its contribution as a representation refinement module. 
First, we compare MPL with existing visual prompt methods within AptM-GCD and analyze the additional gains achieved when combined with AEL. 
Next, to validate MPL as a versatile prompt-based representation learner, we evaluate the generalizability by applying it to other multimodal GCD methods, unimodal GCD methods without textual guidance, and class-imbalanced unimodal GCD settings.

\textbf{\textit{MPL in AptM-GCD.}}
We first evaluate MPL as a purely visual prompting module within AptM-GCD by removing AEL and textual attributes, while keeping multimodal prompt generation and the multimodal prompt transformer. We compare MPL with representative visual prompting methods, including VPT~\cite{vpt}, Global~\cite{global}, and SPM~\cite{spm}. As shown in Fig.~\ref{fig:ab-vp-our}, MPL achieves the best performance among these strategies. Even without AEL, MPL effectively integrates fine-grained visual cues with global context. When combined with AEL, performance improves further, demonstrating that MPL not only acts as a strong visual prompting method, but also provides an effective interface for incorporating attribute-guided local evidence into the visual encoder.

% 用 \providecommand，它的逻辑是：如果没定义过就定义，如果别的表格定义过了就直接复用，绝不报错
\providecommand{\inc}[1]{\raisebox{0.15ex}{\textsuperscript{{\color[HTML]{109e10}\textbf{\tiny +#1}}}}}

\begin{table}[t]
\centering
\begin{threeparttable}
\caption{Transferability of MLP on multimodal GCD methods}
\label{tab:ab-vp-mgcd}
\setlength{\tabcolsep}{2pt}
% 核心安全伞：给每一行天花板塞入 2.5pt 的透明垫片，彻底阻断背景色对上方字体的裁剪
\setlength{\extrarowheight}{2pt}
\newcolumntype{Y}{>{\centering\arraybackslash}X}

\resizebox{\columnwidth}{!}{%
\vbox{%
\begin{tabularx}{\columnwidth}{l YYY YYY}
\toprule
\multirow{2}{*}{\textbf{Method}} & \multicolumn{3}{c}{\textbf{SSB-AVG}} & \multicolumn{3}{c}{\textbf{ImageNet-100}} \\ [2pt] % 给双层表头中间加个缓冲间距
\cmidrule(lr){2-4} \cmidrule(l){5-7}
 & All & Old & New & All & Old & New \\ 
\midrule
TextGCD (ECCV'24) & 71.4 & 71.0 & 71.7 & 91.7 & 95.7 & 89.7 \\ 
\rowcolor[HTML]{E6ECF5} 
\textbf{+ Our MLP} & \textbf{72.6}\inc{1.2} & \textbf{72.0}\inc{1.0} & \textbf{73.5}\inc{1.8} & \textbf{93.2}\inc{1.5} & \textbf{96.1}\inc{0.4} & \textbf{91.9}\inc{2.2} \\ 
\midrule
GET (CVPR'25) & 71.5 & 74.8 & 69.8 & 88.0 & 92.4 & 85.2 \\ 
\rowcolor[HTML]{E6ECF5} 
\textbf{+ Our MLP} & \textbf{73.3}\inc{1.8} & \textbf{74.8}\inc{0.0} & \textbf{72.3}\inc{2.5} & \textbf{90.9}\inc{2.9} & \textbf{92.7}\inc{0.3} & \textbf{88.1}\inc{2.9} \\ 
\midrule
SpectralGCD (ICLR'26) & 77.1 & 79.7 & 75.8 & 93.4 & 96.1 & 92.0 \\ 
\rowcolor[HTML]{E6ECF5} 
\textbf{+ Our MLP} & \textbf{78.3}\inc{1.2} & \textbf{80.1}\inc{0.4} & \textbf{77.4}\inc{1.6} & \textbf{94.3}\inc{0.9} & \textbf{96.1}\inc{0.0} & \textbf{93.5}\inc{1.5} \\ 
\bottomrule
\end{tabularx}%
}}

\begin{tablenotes}[flushleft]
\footnotesize
\item[] ``SSB-Avg'' is the average performance over three fine-grained SSB datasets (CUB \cite{cub}, Stanford Cars \cite{cars}, and Aircraft \cite{aircraft}).
\end{tablenotes}
\end{threeparttable}
% \vspace{-10pt}
\end{table}

\textbf{\textit{MPL in Multimodal GCD.}}
We further evaluate the transferability of MPL as a visual prompting module by incorporating it into recent multimodal GCD methods, including TextGCD~\cite{textgcd}, GET~\cite{get}, and SpectralGCD~\cite{spectralgcd}. As shown in Tab.~\ref{tab:ab-vp-mgcd}, we report results on SSB-Avg, which denotes the average performance over the three fine-grained SSB datasets, and ImageNet-100. MPL consistently improves all methods, regardless of whether they are tag-based or attribute-based. The gains are particularly evident on novel classes, suggesting that prompt-based visual refinement helps existing multimodal GCD frameworks better exploit fine-grained visual cues for discovering unseen categories. These results demonstrate that MPL is not limited to AptM-GCD, but serves as a general visual enhancement module for multimodal GCD.

% 黄金平衡定义：用 \renewcommand 强行覆盖旧定义，改用 \scriptsize 并悬浮 0.5ex
\renewcommand{\inc}[1]{\raisebox{0.5ex}[0pt][0pt]{{\color[HTML]{109e10}\textbf{\scriptsize +#1}}}}

\begin{table}[t]
\centering
\begin{threeparttable}
\caption{Transferability of MLP in unimodal GCD}
\label{tab:ab-vp-gcd}
\setlength{\tabcolsep}{3.3pt} 
\setlength{\extrarowheight}{2pt} % 保持微小的行高缓冲
\newcolumntype{Y}{>{\centering\arraybackslash}X}

\begin{tabularx}{\columnwidth}{@{} l YYY YYY @{} }
\toprule
       & \multicolumn{3}{c}{\textbf{SSB-Avg}} & \multicolumn{3}{c}{\textbf{CIFAR-100}} \\ 
\cmidrule(r){2-4} \cmidrule(l){5-7} 
\multirow{-3}{*}{\textbf{Method}} & All           & Old                    & New                                   & All           & Old           & New           \\ 
\midrule
CMS \cite{cms}  & 57.1          & 65.9                    & 52.7                                  & 79.6           & 83.2          & 72.3          \\ 
\rowcolor[HTML]{E6ECF5} 
\textbf{+ Our MLP}       & \textbf{60.5}\inc{3.4} & \textbf{68.3}\inc{2.4}           & \textbf{57.5}\inc{4.8}                       & \textbf{82.0}\inc{2.4} & \textbf{84.2}\inc{1.0} & \textbf{76.6}\inc{4.3} \\ 
\midrule
CiPR \cite{cipr}  & 52.1          & 60.1                    & 47.9                                  & 81.5          & 82.4          & 79.7          \\ 
\rowcolor[HTML]{E6ECF5} 
\textbf{+ Our MLP}       & \textbf{56.2}\inc{4.1} & \textbf{63.9}\inc{3.8}           & \textbf{53.1}\inc{5.2}                       & \textbf{81.9}\inc{0.4} & \textbf{82.5}\inc{0.1}          & \textbf{81.9}\inc{2.2} \\ 
\midrule
LegoGCD \cite{legogcd} & 58.7          & 69.7                    & 53.3                                  & 81.8          & 81.4          & 82.5          \\ 
\rowcolor[HTML]{E6ECF5} 
\textbf{+ Our MLP}       & \textbf{64.5}\inc{5.8} & \textbf{73.1}\inc{3.4}           & \textbf{60.6}\inc{7.3}                      & \textbf{82.8}\inc{0.4} & \textbf{81.8}\inc{0.4}           & \textbf{85.5}\inc{3.0} \\ 
\bottomrule
\end{tabularx}

\begin{tablenotes}[flushleft]
\footnotesize
\item[] We integrate MLP into several advanced GCD methods, including CMS (CVPR'24), CiPR (TMLR'24), and LegoGCD (CVPR'24), to evaluate the transferability.
\end{tablenotes}
\end{threeparttable}
\vspace{-10pt}
\end{table}

\textbf{\textit{MPL in Unimodal GCD.}}
To examine the generalization and robustness of MPL without textual guidance, we further evaluate it as a vision prompt learning technique in unimodal GCD. MPL is integrated into recent unimodal GCD methods, including CMS~\cite{cms}, CiPR~\cite{cipr}, and LegoGCD~\cite{legogcd}. As shown in Tab.~\ref{tab:ab-vp-gcd}, MPL consistently improves both clustering-based~\cite{cms,cipr} and parametric-classifier-based~\cite{legogcd} methods, with particularly notable gains on novel classes. For instance, on SSB-Avg, MPL increases the New accuracy of CMS, CiPR, and LegoGCD by 4.8\%, 5.2\%, and 7.3\%, respectively. Since MPL refines visual representations through prompt generation and re-injection within the visual encoder, it preserves the original GCD objective and classifier design while introducing lightweight prompt-based refinement modules. These results demonstrate that MPL is a general and effective prompting strategy that can be applied to both unimodal visual prompting and multimodal attribute-guided settings.

\begin{table}[t]
\centering 
\begin{threeparttable}
% 直接在标题中注明是 CIFAR-10-LT
\caption{Transferability of MLP under class-imbalanced GCD on \textbf{CIFAR-10-LT}}
\label{tab:ab-vp-ungcd}
\setlength{\tabcolsep}{3.3pt} 
\setlength{\extrarowheight}{3pt} 
\newcolumntype{Y}{>{\centering\arraybackslash}X}

\begin{tabularx}{\columnwidth}{@{} l c YYY YYY @{}}
\toprule
 & & \multicolumn{3}{c}{\textbf{$p=20$}} & \multicolumn{3}{c}{\textbf{$p=100$}} \\ 
\cmidrule(lr){3-5} \cmidrule(l){6-8}  
\textbf{Method} & \textbf{Type} & New & Old & All & New & Old & All \\ 
\midrule
SimGCD \cite{simgcd} (ICCV-23) & Standard GCD & 78.9 & 77.4 & 77.6 & 55.9 & 77.4 & 67.9 \\
\rowcolor[HTML]{E6ECF5} 
\textbf{+ Our MLP} & & \textbf{85.0} & \textbf{83.2} & \textbf{84.1} & \textbf{73.1} & \textbf{81.6} & \textbf{77.4} \\ 
\midrule
CMS \cite{cms} (CVPR-24) & Standard GCD & 75.0 & 84.6 & 78.8 & 66.4 & 71.6 & 69.1 \\
\rowcolor[HTML]{E6ECF5} 
\textbf{+ Our MLP} & & \textbf{84.1} & \textbf{86.8} & \textbf{86.6} & \textbf{83.0} & \textbf{74.5} & \textbf{78.8} \\
\bottomrule
\end{tabularx}

\begin{tablenotes}[flushleft]
\footnotesize
\item[] The imbalance ratio $p$ denotes the ratio between the largest and smallest class sizes, with $p=20$ and $p=100$ evaluated.
\end{tablenotes}
\end{threeparttable}
\end{table}

\textbf{\textit{MPL in Class-Imbalanced GCD.}}
We evaluate the robustness of MPL under long-tailed data distributions. Following~\cite{bacon}, we construct CIFAR-10-LT from CIFAR-10, where the imbalance ratio $p$ is defined as the number of samples in the largest class divided by that in the smallest class. Two imbalance ratios are considered to evaluate performance under different degrees of class imbalance, with the unlabeled set containing samples from both old and novel classes.
As shown in Tab.~\ref{tab:ab-vp-ungcd}, integrating MPL into standard GCD methods consistently improves performance under both imbalance settings. In particular, SimGCD+Ours and CMS+Ours outperform their original counterparts, with substantial gains on novel classes. Under the more challenging setting of $p=100$, the novel-class accuracy of SimGCD increases from 55.9\% to 73.1\%, and CMS improves from 66.4\% to 83.0\%. These results demonstrate that MPL is robust to class imbalance and effectively enhances category discovery for under-represented novel classes.

\begin{table}[t]
\centering
\begin{threeparttable}
\caption{Effect of the feature fuse in multimodal prompt generator.}
\label{tab:ab-fuse}
% 1. 使用 tabcolsep 微调列间距
\setlength{\tabcolsep}{4.5pt} 
% 2. 保证带有 rowcolor 的行上下有微小的呼吸空间
\setlength{\extrarowheight}{2pt} 
% 3. 定义一个自动居中且自适应宽度的列类型 Y
\newcolumntype{Y}{>{\centering\arraybackslash}X}

% 4. 彻底扔掉 \resizebox，改用 tabularx
\begin{tabularx}{\columnwidth}{@{} l YYYYYYYYY @{}}
\toprule
 & \multicolumn{3}{c}{\textbf{CUB}} & \multicolumn{3}{c}{\textbf{Stanford Cars}} & \multicolumn{3}{c}{\textbf{Aircraft}} \\ 
\cmidrule(r){2-4} \cmidrule(lr){5-7} \cmidrule(l){8-10}
\multirow{-3}{*}{\textbf{Fuse}} & All & Old & New & All & Old & New & All & Old & New \\ 
\midrule
\ding{55} & 80.5 & 80.1 & 80.8 & 90.3 & 93.6 & 88.4 & 64.6 & 64.6 & 64.6 \\ 
\rowcolor[HTML]{E6ECF5} 
\ding{51} & \textbf{81.5} & \textbf{80.6} & \textbf{81.9} & \textbf{91.0} & \textbf{94.0} & \textbf{89.2} & \textbf{65.3} & \textbf{65.5} & \textbf{65.3} \\ 
\bottomrule
\end{tabularx}

\begin{tablenotes}[flushleft]
\footnotesize
\item[] Fuse (\ding{55}) denotes directly utilizing $F_{\text{attr}}$ as the prompt input.
\end{tablenotes}
\end{threeparttable}
\vspace{-10pt}
\end{table}

\subsubsection{\textbf{Effect of Feature Fusion in Multimodal Prompt Generation}}
We further analyze the effect of feature fusion in the multimodal prompt generator, as shown in Table~\ref{tab:ab-fuse}. Using only the attribute-guided feature map \(\mathbf{F}^{\mathrm{attr}}\) leads to a slight performance drop, indicating that attribute evidence alone is not sufficient for stable and comprehensive prompt generation.
It is worth noting that \(\mathbf{F}^{\mathrm{attr}}\) is obtained by residual attribute-guided recalibration rather than hard filtering, so the original visual structure is largely preserved. Thus, \(\mathbf{F}^{p}\) and \(\mathbf{F}^{\mathrm{attr}}\) are complementary rather than redundant: the former retains complete visual context, while the latter emphasizes localized discriminative evidence.
By fusing \(\mathbf{F}^{\mathrm{attr}}\) with the original patch feature map \(\mathbf{F}^{p}\), the prompt generator can jointly access the complete visual context and the locally discriminative evidence emphasized by attributes.
Moreover, the learnable fusion convolution adaptively balances the two inputs, allowing the model to decide how much original context and attribute-enhanced evidence should contribute to prompt generation. This is especially important during early training, since attribute selection and grounding may still be noisy. Relying solely on \(\mathbf{F}^{\mathrm{attr}}\) can make the prompt overly dependent on incomplete local evidence.

\begin{figure}[t]
\centering
\includegraphics[width=0.8\linewidth]{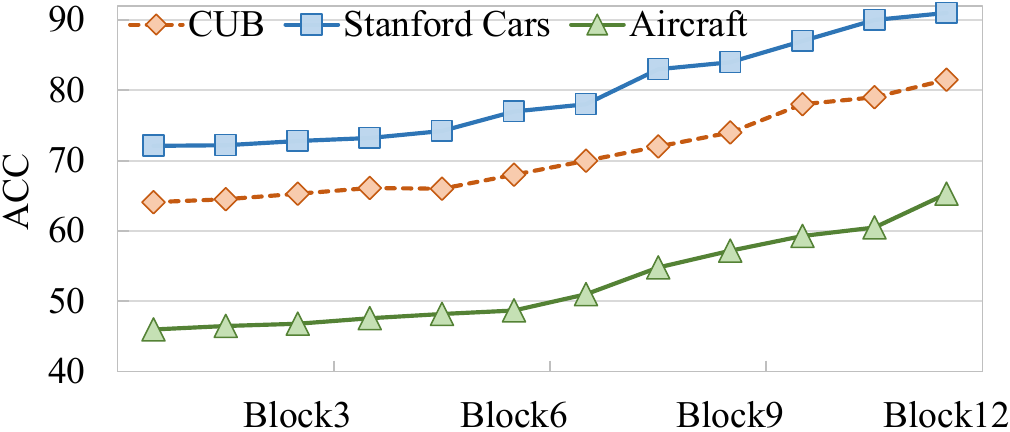}
\caption{\textbf{Effect of prompt placement in the Multimodal Prompt Transformer.} We inject multimodal prompts into different transformer blocks to study layer-wise performance on three fine-grained datasets. }
\label{fig:ab-mpt-pp}
\vspace{-10pt}
\end{figure}

\subsubsection{\textbf{Effect of Prompt Placement in Multimodal Prompt Transformer}}
We investigate the layer-wise placement of prompts in the Multimodal Prompt Transformer (MPT) by injecting them into different transformer blocks on three fine-grained datasets. As shown in Fig.~\ref{fig:ab-mpt-pp}, ``All'' performance improves when prompts are inserted into deeper blocks, while early insertion can even degrade accuracy. This indicates that prompts are most effective when applied to high-level semantic features rather than low-level visual patterns. Deeper layers encode more abstract, category-relevant representations, making prompt-guided refinement at these stages particularly beneficial, consistent with prior observations~\cite{pt-finding}.

\begin{table}[t]
\centering
\caption{Ablation study of training strategy of Prompt Transformer and Prompt Generator on the SSB and ImageNet.}
\label{tab:ab-mpl-train}
\setlength{\tabcolsep}{7pt} % 稍微缩小列间距以完美适应两组数据集
\resizebox{\columnwidth}{!}{%
\begin{tabular}{l ccc ccc}
\toprule
 & \multicolumn{3}{c}{\textbf{SSB-Avg}} & \multicolumn{3}{c}{\textbf{ImageNet-100}} \\ 
\cmidrule(r){2-4} \cmidrule(l){5-7} % 在中间留出美观的视觉间隙
\multirow{-3}{*}{\textbf{Training Strategy}} & All           & Old           & New           & All           & Old           & New           \\ \midrule
Decoupling        & 77.7          & 79.8         & 76.2          & 93.6          & 95.8          & 92.6          \\
Alternating       & 78.1          & \textbf{80.4}          & 77.8          & 94.1          & 96.1          & 93.0          \\
\rowcolor[HTML]{E6ECF5} 
Joint             & \textbf{79.3} & 78.4 & \textbf{78.8} & \textbf{95.1} & \textbf{96.8} & \textbf{94.4} \\ \bottomrule
\end{tabular}%
}
\vspace{-10pt}
\end{table}

\begin{table}[t]
\centering
\caption{Comparison of model size with multimodal GCD methods.}
\label{tab:param}
\setlength{\tabcolsep}{3pt} 
\setlength{\extrarowheight}{2pt} 
\resizebox{\columnwidth}{!}{
\begin{tabular}{lccc}
\hline
\textbf{Method} & \textbf{Backbone} & \textbf{Extra Modules} & \textbf{Total Params (M)} \\
\hline
TextGCD \cite{textgcd}      & CLIP ViT-B/16 & classifiers & $\sim$150 \\
SpectralGCD \cite{spectralgcd}  & CLIP ViT-B/16 & Projection   & $\sim$156 \\
AptM-GCD     & CLIP ViT-B/16 & Projection + AEL + MPL                 & $\sim$170 \\
\hline
\end{tabular}
}
\end{table}

\subsubsection{\textbf{Effect of Training Strategy}}
We study different strategies for training the multimodal prompt generator (MPG) and multimodal prompt transformer (MPT) in MPL. Tab.~\ref{tab:ab-mpl-train} compares three approaches: decoupled, alternating, and joint training. Decoupled training optimizes MPG and MPT separately, limiting their collaboration. Alternating training allows interaction but may suffer temporal misalignment, as MPG prompts may not always match MPT’s current state. In contrast, joint end-to-end training synchronizes prompt generation and prompt-guided attention, enabling mutual reinforcement between the two modules and yielding the best performance.

\subsubsection{\textbf{Parameter Analysis}}
We further compare the model size of AptM-GCD with representative multimodal GCD methods. As shown in Tab. \ref{tab:param}, the proposed MPL introduces approximately 15M additional parameters, resulting in a total model size of about 170M parameters.  Under the same CLIP ViT-B/16 backbone, TextGCD \cite{textgcd} and SpectralGCD \cite{spectralgcd} contain approximately 150M and 156M parameters, respectively. Therefore, AptM-GCD incurs a moderate increase in model size in exchange for consistent performance gains across different benchmarks.

\section{Conclusion}
In this paper, we presented AptM-GCD, a multimodal framework for harnessing fine-grained attributes as apt textual guidance in generalized category discovery. We first identified that directly treating attributes as global textual cues can introduce spurious semantic associations and mislead category discovery, especially harming novel-class discovery. To address this, AptM-GCD grounds attributes into object-aware local visual evidence via Attribute-Evidence Learning, and then injects this evidence into the visual transformer through Multimodal Prompt Learning for context-aware representation refinement. Extensive experiments on six benchmarks demonstrate that AptM-GCD achieves state-of-the-art performance. Moreover, MPL shows strong transferability to existing multimodal methods, unimodal GCD, and class-imbalanced settings. Our results highlight that the true value of attributes lies not in adding more textual cues, but in transforming them into apt guidance for category discovery.
Nevertheless, challenging cases with insufficient visual evidence, such as background-camouflaged and heavily occluded objects, remain difficult for GCD due to unreliable visual-semantic correspondence. Future work will explore more robust object-centric representations and compositional visual-semantic reasoning to further advance open-world category discovery.

\bibliographystyle{IEEEtran}
\bibliography{tpami2026}

@String(CVPR  = {IEEE Conf. Comput. Vis. Pattern Recog.})

@String(ICLR  = {Int. Conf. Learn. Represent.})

@String(AAAI  = {AAAI})

@String(CVPR  = {CVPR})

@String(ICLR  = {ICLR})

@article{humannovel1,
  title={Novel class discovery: an introduction and key concepts},
  author={Troisemaine, Colin and Lemaire, Vincent and Gosselin, St{\'e}phane and Reiffers-Masson, Alexandre and Flocon-Cholet, Joachim and Vaton, Sandrine},
  journal={arXiv preprint arXiv:2302.12028},
  year={2023}
}

@article{humannovel2,
  title={Autonovel: Automatically discovering and learning novel visual categories},
  author={Han, Kai and Rebuffi, Sylvestre-Alvise and Ehrhardt, Sebastien and Vedaldi, Andrea and Zisserman, Andrew},
  journal={IEEE Transactions on Pattern Analysis and Machine Intelligence},
  volume={44},
  number={10},
  pages={6767--6781},
  year={2021},
  publisher={IEEE}
}

@article{deeplearning1,
  title={Open-world machine learning: A review and new outlooks},
  author={Zhu, Fei and Ma, Shijie and Cheng, Zhen and Zhang, Xu-Yao and Zhang, Zhaoxiang and Tao, Dacheng and Liu, Cheng-Lin},
  journal={arXiv preprint arXiv:2403.01759},
  year={2024}
}

@article{deeplearning2,
  title={Recent advances in open set recognition: A survey},
  author={Geng, Chuanxing and Huang, Sheng-jun and Chen, Songcan},
  journal={IEEE transactions on pattern analysis and machine intelligence},
  volume={43},
  number={10},
  pages={3614--3631},
  year={2020},
  publisher={IEEE}
}

@article{deeplearning3,
  title={A unified survey on anomaly, novelty, open-set, and out-of-distribution detection: Solutions and future challenges},
  author={Salehi, Mohammadreza and Mirzaei, Hossein and Hendrycks, Dan and Li, Yixuan and Rohban, Mohammad Hossein and Sabokrou, Mohammad},
  journal={arXiv preprint arXiv:2110.14051},
  year={2021}
}

@inproceedings{gcd,
  title={Generalized category discovery},
  author={Vaze, Sagar and Han, Kai and Vedaldi, Andrea and Zisserman, Andrew},
  booktitle={Proceedings of the IEEE/CVF conference on computer vision and pattern recognition},
  pages={7492--7501},
  year={2022}
}

@inproceedings{dccl,
  title={Dynamic conceptional contrastive learning for generalized category discovery},
  author={Pu, Nan and Zhong, Zhun and Sebe, Nicu},
  booktitle={Proceedings of the IEEE/CVF conference on computer vision and pattern recognition},
  pages={7579--7588},
  year={2023}
}

@inproceedings{simgcd,
  title={Parametric classification for generalized category discovery: A baseline study},
  author={Wen, Xin and Zhao, Bingchen and Qi, Xiaojuan},
  booktitle={Proceedings of the IEEE/CVF international conference on computer vision},
  pages={16590--16600},
  year={2023}
}

@article{ncd1,
  title={Automatically discovering and learning new visual categories with ranking statistics},
  author={Han, Kai and Rebuffi, Sylvestre-Alvise and Ehrhardt, Sebastien and Vedaldi, Andrea and Zisserman, Andrew},
  journal={arXiv preprint arXiv:2002.05714},
  year={2020}
}

@inproceedings{ncd2,
  title={Modeling inter-class and intra-class constraints in novel class discovery},
  author={Li, Wenbin and Fan, Zhichen and Huo, Jing and Gao, Yang},
  booktitle={Proceedings of the IEEE/CVF conference on computer vision and pattern recognition},
  pages={3449--3458},
  year={2023}
}

@article{ncd3,
  title={Class-relation knowledge distillation for novel class discovery},
  author={Gu, Peiyan and Zhang, Chuyu and Xu, Ruijie and He, Xuming},
  journal={lamp},
  volume={12},
  number={15.0},
  pages={17--5},
  year={2023}
}

@article{osr1,
  title={Adversarial reciprocal points learning for open set recognition},
  author={Chen, Guangyao and Peng, Peixi and Wang, Xiangqian and Tian, Yonghong},
  journal={IEEE Transactions on Pattern Analysis and Machine Intelligence},
  volume={44},
  number={11},
  pages={8065--8081},
  year={2021},
  publisher={IEEE}
}

@article{osr2,
  title={Open long-tailed recognition in a dynamic world},
  author={Liu, Ziwei and Miao, Zhongqi and Zhan, Xiaohang and Wang, Jiayun and Gong, Boqing and Yu, Stella X},
  journal={IEEE Transactions on Pattern Analysis and Machine Intelligence},
  volume={46},
  number={3},
  pages={1836--1851},
  year={2022},
  publisher={IEEE}
}

@article{humanmultimodal,
  title={From perceptual categories to concepts: What develops?},
  author={Sloutsky, Vladimir M},
  journal={Cognitive science},
  volume={34},
  number={7},
  pages={1244--1286},
  year={2010},
  publisher={Wiley Online Library}
}

@inproceedings{textgcd,
  title={Textual knowledge matters: Cross-modality co-teaching for generalized visual class discovery},
  author={Zheng, Haiyang and Pu, Nan and Li, Wenjing and Sebe, Nicu and Zhong, Zhun},
  booktitle={European Conference on Computer Vision},
  pages={41--58},
  year={2024},
  organization={Springer}
}

@article{spectralgcd,
  title={SpectralGCD: Spectral concept selection and cross-modal representation learning for generalized category discovery},
  author={Caselli, Lorenzo and Mistretta, Marco and Magistri, Simone and Bagdanov, Andrew D},
  journal={arXiv preprint arXiv:2602.17395},
  year={2026}
}

@inproceedings{legogcd,
  title={Solving the catastrophic forgetting problem in generalized category discovery},
  author={Cao, Xinzi and Zheng, Xiawu and Wang, Guanhong and Yu, Weijiang and Shen, Yunhang and Li, Ke and Lu, Yutong and Tian, Yonghong},
  booktitle={Proceedings of the IEEE/CVF Conference on Computer Vision and Pattern Recognition},
  pages={16880--16889},
  year={2024}
}

@article{sptnet,
  title={Sptnet: An efficient alternative framework for generalized category discovery with spatial prompt tuning},
  author={Wang, Hongjun and Vaze, Sagar and Han, Kai},
  journal={arXiv preprint arXiv:2403.13684},
  year={2024}
}

@article{protogcd,
  title={ProtoGCD: Unified and unbiased prototype learning for generalized category discovery},
  author={Ma, Shijie and Zhu, Fei and Zhang, Xu-Yao and Liu, Cheng-Lin},
  journal={IEEE Transactions on Pattern Analysis and Machine Intelligence},
  year={2025},
  publisher={IEEE}
}

@article{meanteacher,
  title={Mean teachers are better role models: Weight-averaged consistency targets improve semi-supervised deep learning results},
  author={Tarvainen, Antti and Valpola, Harri},
  journal={Advances in neural information processing systems},
  volume={30},
  year={2017}
}

@inproceedings{selfdis1,
  title={Masked siamese networks for label-efficient learning},
  author={Assran, Mahmoud and Caron, Mathilde and Misra, Ishan and Bojanowski, Piotr and Bordes, Florian and Vincent, Pascal and Joulin, Armand and Rabbat, Mike and Ballas, Nicolas},
  booktitle={European conference on computer vision},
  pages={456--473},
  year={2022},
  organization={Springer}
}

@article{ssr2gcd,
  title={Multi-Modal Representation Learning via Semi-Supervised Rate Reduction for Generalized Category Discovery},
  author={He, Wei and Meng, Xianghan and Huang, Zhiyuan and Qi, Xianbiao and Xiao, Rong and Li, Chun-Guang},
  journal={arXiv preprint arXiv:2602.19910},
  year={2026}
}

@inproceedings{dino1,
  title={Emerging properties in self-supervised vision transformers},
  author={Caron, Mathilde and Touvron, Hugo and Misra, Ishan and J{\'e}gou, Herv{\'e} and Mairal, Julien and Bojanowski, Piotr and Joulin, Armand},
  booktitle={Proceedings of the IEEE/CVF international conference on computer vision},
  pages={9650--9660},
  year={2021}
}

@article{minmax,
  title={39 Dimensionality and sample size considerations in pattern recognition practice},
  author={Jain, Anil K and Chandrasekaran, Balakrishnan},
  journal={Handbook of statistics},
  volume={2},
  pages={835--855},
  year={1982},
  publisher={Elsevier}
}

@inproceedings{object2,
  title={Self-supervised transformers for unsupervised object discovery using normalized cut},
  author={Wang, Yangtao and Shen, Xi and Hu, Shell Xu and Yuan, Yuan and Crowley, James L and Vaufreydaz, Dominique},
  booktitle={Proceedings of the IEEE/CVF Conference on Computer Vision and Pattern Recognition},
  pages={14543--14553},
  year={2022}
}

@article{crossattn1,
  title={Attention is all you need},
  author={Vaswani, Ashish and Shazeer, Noam and Parmar, Niki and Uszkoreit, Jakob and Jones, Llion and Gomez, Aidan N and Kaiser, {\L}ukasz and Polosukhin, Illia},
  journal={Advances in neural information processing systems},
  volume={30},
  year={2017}
}

@inproceedings{crossattn2,
  title={Mdetr-modulated detection for end-to-end multi-modal understanding},
  author={Kamath, Aishwarya and Singh, Mannat and LeCun, Yann and Synnaeve, Gabriel and Misra, Ishan and Carion, Nicolas},
  booktitle={Proceedings of the IEEE/CVF international conference on computer vision},
  pages={1780--1790},
  year={2021}
}

@article{TransZero++,
  author       = {Shiming Chen and
                  Ziming Hong and
                  Wenjin Hou and
                  Guo{-}Sen Xie and
                  Yibing Song and
                  Jian Zhao and
                  Xinge You and
                  Shuicheng Yan and
                  Ling Shao},
  title        = {TransZero++: Cross Attribute-Guided Transformer for Zero-Shot Learning},
  journal      = {IEEE transactions on pattern analysis and machine intelligence},
  volume       = {45},
  number       = {11},
  pages        = {12844--12861},
  year         = {2023}
}

@inproceedings{aptgcd,
  title={Less attention is more: Prompt transformer for generalized category discovery},
  author={Zhang, Wei and Zhang, Baopeng and Teng, Zhu and Luo, Wenxin and Zou, Junnan and Fan, Jianping},
  booktitle={Proceedings of the Computer Vision and Pattern Recognition Conference},
  pages={30322--30331},
  year={2025}
}

@inproceedings{metalearning,
  title={Model-agnostic meta-learning for fast adaptation of deep networks},
  author={Finn, Chelsea and Abbeel, Pieter and Levine, Sergey},
  booktitle={International conference on machine learning},
  pages={1126--1135},
  year={2017},
  organization={PMLR}
}

@inproceedings{glu,
  title={Language modeling with gated convolutional networks},
  author={Dauphin, Yann N and Fan, Angela and Auli, Michael and Grangier, David},
  booktitle={International conference on machine learning},
  pages={933--941},
  year={2017},
  organization={PMLR}
}

@inproceedings{deepwise,
  title={Deep learning with depth wise separable convolutions},
  author={Fran, Chollet and others},
  booktitle={IEEE conference on computer vision and pattern recognition (CVPR)},
  volume={78},
  year={2017}
}

@article{prefix,
  title={Towards a unified view on visual parameter-efficient transfer learning},
  author={Bruce, XB and Chang, Jianlong and Liu, Lingbo and Tian, Qi and Chen, Chang Wen},
  journal={CoRR},
  year={2022}
}

@inproceedings{crossattn3,
  title={Multimodal Fusion with Cross-Modal Attention for Action Recognition in Still Images},
  author={Tsai, Jia-Hua and Chu, Wei-Ta},
  booktitle={Proceedings of the 4th ACM International Conference on Multimedia in Asia},
  pages={1--5},
  year={2022}
}

@article{clipgcd,
  title={Clip-gcd: Simple language guided generalized category discovery},
  author={Ouldnoughi, Rabah and Kuo, Chia-Wen and Kira, Zsolt},
  journal={arXiv preprint arXiv:2305.10420},
  year={2023}
}

@inproceedings{promptcal,
  title={Promptcal: Contrastive affinity learning via auxiliary prompts for generalized novel category discovery},
  author={Zhang, Sheng and Khan, Salman and Shen, Zhiqiang and Naseer, Muzammal and Chen, Guangyi and Khan, Fahad Shahbaz},
  booktitle={Proceedings of the IEEE/CVF conference on computer vision and pattern recognition},
  pages={3479--3488},
  year={2023}
}

@inproceedings{cms,
  title={Contrastive mean-shift learning for generalized category discovery},
  author={Choi, Sua and Kang, Dahyun and Cho, Minsu},
  booktitle={Proceedings of the IEEE/CVF conference on computer vision and pattern recognition},
  pages={23094--23104},
  year={2024}
}

@article{cipr,
  title={Cipr: An efficient framework with cross-instance positive relations for generalized category discovery},
  author={Hao, Shaozhe and Han, Kai and Wong, Kwan-Yee K},
  journal={arXiv preprint arXiv:2304.06928},
  year={2023}
}

@inproceedings{allgcd,
  title={AllGCD: Leveraging All Unlabeled Data for Generalized Category Discovery},
  author={Cao, Xinzi and Chen, Ke and Yang, Feidiao and Zheng, Xiawu and Tian, Yonghong and Lu, Yutong},
  booktitle={Proceedings of the IEEE/CVF International Conference on Computer Vision},
  pages={3293--3303},
  year={2025}
}

@inproceedings{get,
  title={Get: Unlocking the multi-modal potential of clip for generalized category discovery},
  author={Wang, Enguang and Peng, Zhimao and Xie, Zhengyuan and Yang, Fei and Liu, Xialei and Cheng, Ming-Ming},
  booktitle={Proceedings of the Computer Vision and Pattern Recognition Conference},
  pages={20296--20306},
  year={2025}
}

@article{coop,
  title={Learning to prompt for vision-language models},
  author={Zhou, Kaiyang and Yang, Jingkang and Loy, Chen Change and Liu, Ziwei},
  journal={International journal of computer vision},
  volume={130},
  number={9},
  pages={2337--2348},
  year={2022},
  publisher={Springer}
}

@inproceedings{cocoop,
  title={Conditional prompt learning for vision-language models},
  author={Zhou, Kaiyang and Yang, Jingkang and Loy, Chen Change and Liu, Ziwei},
  booktitle={Proceedings of the IEEE/CVF conference on computer vision and pattern recognition},
  pages={16816--16825},
  year={2022}
}

@inproceedings{vpt,
  title={Visual prompt tuning},
  author={Jia, Menglin and Tang, Luming and Chen, Bor-Chun and Cardie, Claire and Belongie, Serge and Hariharan, Bharath and Lim, Ser-Nam},
  booktitle={European conference on computer vision},
  pages={709--727},
  year={2022},
  organization={Springer}
}

@inproceedings{maple,
  title={Maple: Multi-modal prompt learning},
  author={Khattak, Muhammad Uzair and Rasheed, Hanoona and Maaz, Muhammad and Khan, Salman and Khan, Fahad Shahbaz},
  booktitle={Proceedings of the IEEE/CVF conference on computer vision and pattern recognition},
  pages={19113--19122},
  year={2023}
}

@inproceedings{promptsrc,
  title={Self-regulating prompts: Foundational model adaptation without forgetting},
  author={Khattak, Muhammad Uzair and Wasim, Syed Talal and Naseer, Muzammal and Khan, Salman and Yang, Ming-Hsuan and Khan, Fahad Shahbaz},
  booktitle={Proceedings of the IEEE/CVF international conference on computer vision},
  pages={15190--15200},
  year={2023}
}

@inproceedings{argue,
  title={Argue: Attribute-guided prompt tuning for vision-language models},
  author={Tian, Xinyu and Zou, Shu and Yang, Zhaoyuan and Zhang, Jing},
  booktitle={Proceedings of the IEEE/CVF Conference on Computer Vision and Pattern Recognition},
  pages={28578--28587},
  year={2024}
}

@inproceedings{atprompt,
  title={Advancing textual prompt learning with anchored attributes},
  author={Li, Zheng and Song, Yibing and Cheng, Ming-Ming and Li, Xiang and Yang, Jian},
  booktitle={Proceedings of the IEEE/CVF International Conference on Computer Vision},
  pages={3618--3627},
  year={2025}
}

@article{cub,
  title={The caltech-ucsd birds-200-2011 dataset},
  author={Wah, Catherine and Branson, Steve and Welinder, Peter and Perona, Pietro and Belongie, Serge},
  year={2011},
  publisher={California Institute of Technology}
}

@inproceedings{cars,
  author       = {Jonathan Krause and
                  Michael Stark and
                  Jia Deng and
                  Li Fei{-}Fei},
  title        = {3D Object Representations for Fine-Grained Categorization},
  booktitle    = {Proceedings of the IEEE/CVF international conference on computer vision},
  pages        = {554--561},
  publisher    = {{IEEE} Computer Society},
  year         = {2013},
  doi          = {10.1109/ICCVW.2013.77}
}

@inproceedings{ssb,
  author       = {Sagar Vaze and
                  Kai Han and
                  Andrea Vedaldi and
                  Andrew Zisserman},
  title        = {Open-Set Recognition: {A} Good Closed-Set Classifier is All You Need},
  booktitle    = {International Conference on Learning Representations (ICLR)},
  publisher    = {OpenReview.net},
  year         = {2022},
  bibsource    = {dblp computer science bibliography, https://dblp.org}
}

@misc{aircraft,
  author       = {Subhransu Maji and
                  Esa Rahtu and
                  Juho Kannala and
                  Matthew B. Blaschko and
                  Andrea Vedaldi},
  title        = {Fine-Grained Visual Classification of Aircraft},
  journal      = {CoRR},
  volume       = {abs/1306.5151},
  year         = {2013},
  url          = {http://arxiv.org/abs/1306.5151},
  archivePrefix    = "arXiv",
  eprint       = {1306.5151},
  bibsource    = {dblp computer science bibliography, https://dblp.org}
}

@article{cifar,
  title={Learning multiple layers of features from tiny images},
  author={Krizhevsky, Alex and Hinton, Geoffrey and others},
  year={2009},
  publisher={Toronto, ON, Canada}
}

@inproceedings{imagenet,
  title={Imagenet: A large-scale hierarchical image database},
  author={Deng, Jia and Dong, Wei and Socher, Richard and Li, Li-Jia and Li, Kai and Fei-Fei, Li},
  booktitle={2009 IEEE conference on computer vision and pattern recognition},
  pages={248--255},
  year={2009},
  organization={Ieee}
}

@article{hungarian,
  title={The Hungarian method for the assignment problem},
  author={Kuhn, Harold W},
  journal={Naval research logistics quarterly},
  volume={2},
  number={1-2},
  pages={83--97},
  year={1955},
  publisher={Wiley Online Library}
}

@inproceedings{clip,
  title={Learning transferable visual models from natural language supervision},
  author={Radford, Alec and Kim, Jong Wook and Hallacy, Chris and Ramesh, Aditya and Goh, Gabriel and Agarwal, Sandhini and Sastry, Girish and Askell, Amanda and Mishkin, Pamela and Clark, Jack and others},
  booktitle={International conference on machine learning},
  pages={8748--8763},
  year={2021},
  organization={PmLR}
}

@article{global,
  title={Exploring visual prompts for adapting large-scale models},
  author={Bahng, Hyojin and Jahanian, Ali and Sankaranarayanan, Swami and Isola, Phillip},
  journal={arXiv preprint arXiv:2203.17274},
  year={2022}
}

@inproceedings{spm,
  title={Self-prompt mechanism for few-shot image recognition},
  author={Song, Mingchen and Wang, Huiqiang and Zhong, Guoqiang},
  booktitle={Proceedings of the AAAI Conference on Artificial Intelligence},
  volume={38},
  number={5},
  pages={4934--4942},
  year={2024}
}

@article{bacon,
  title={Towards distribution-agnostic generalized category discovery},
  author={Bai, Jianhong and Liu, Zuozhu and Wang, Hualiang and Chen, Ruizhe and Mu, Lianrui and Li, Xiaomeng and Zhou, Joey Tianyi and Feng, Yang and Wu, Jian and Hu, Haoji},
  journal={Advances in Neural Information Processing Systems},
  volume={36},
  pages={58625--58647},
  year={2023}
}

@inproceedings{pt-finding,
  title={Strong baselines for parameter-efficient few-shot fine-tuning},
  author={Basu, Samyadeep and Hu, Shell and Massiceti, Daniela and Feizi, Soheil},
  booktitle={Proceedings of the AAAI Conference on Artificial Intelligence},
  volume={38},
  number={10},
  pages={11024--11031},
  year={2024}
}

@article{u2net,
  title={U2-Net: Going deeper with nested U-structure for salient object detection},
  author={Qin, Xuebin and Zhang, Zichen and Huang, Chenyang and Dehghan, Masood and Zaiane, Osmar R and Jagersand, Martin},
  journal={Pattern recognition},
  volume={106},
  pages={107404},
  year={2020},
  publisher={Elsevier}
}

@article{dong1,
  title={Selective cross-view topology for deep incomplete multi-view clustering},
  author={Dong, Zhibin and Hu, Dayu and Jin, Jiaqi and Wang, Siwei and Liu, Xinwang and Zhu, En},
  journal={IEEE Transactions on Image Processing},
  volume={34},
  pages={4792--4805},
  year={2025},
  publisher={IEEE}
}

@article{dong2,
  title={Cross-view graph matching for unsupervised learning with unaligned multi-view clustering},
  author={Dong, Zhibin and Yu, Shengju and Wang, Siwei and Liu, Xinwang and Zhu, En},
  journal={IEEE Transactions on Knowledge and Data Engineering},
  year={2026},
  publisher={IEEE}
}

% \newpage

\section{Biography Section}
\vspace{-40pt}

\begin{IEEEbiography}[{\includegraphics[width=1in,height=1.25in,clip,keepaspectratio]{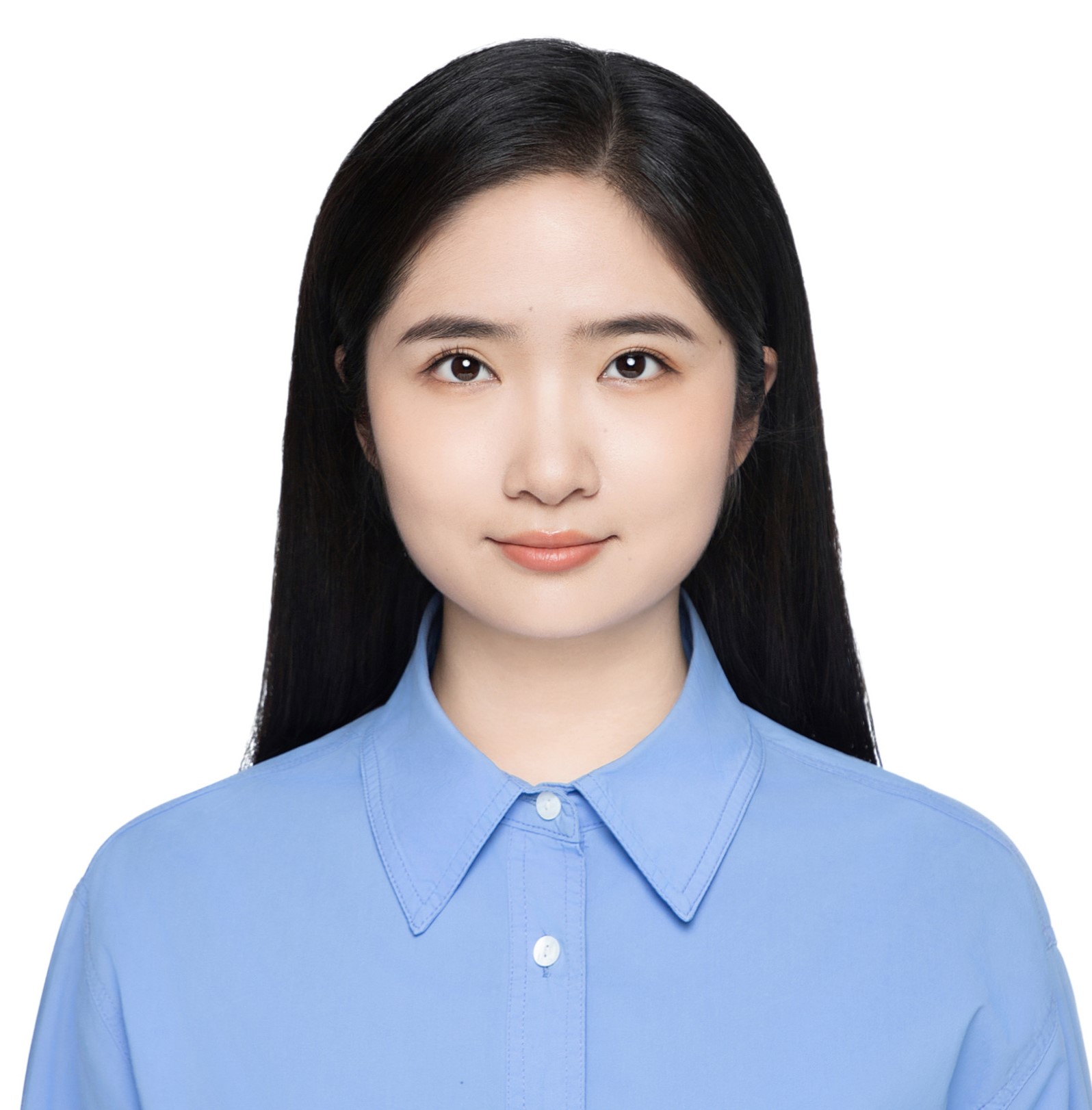}}]{Wei Zhang}
received the M.S. degree in computer science and technology from the University of Science and Technology Beijing, Beijing, China, in 2023. She is currently pursuing the Ph.D. degree in computer science and technology at Beijing Jiaotong University, Beijing, China. Her current research interests include computer vision, multimodal open-world visual perception, and deep learning.
\vspace{-40pt}
\end{IEEEbiography}

\begin{IEEEbiography}[{\includegraphics[width=1in,height=1.25in,clip,keepaspectratio]{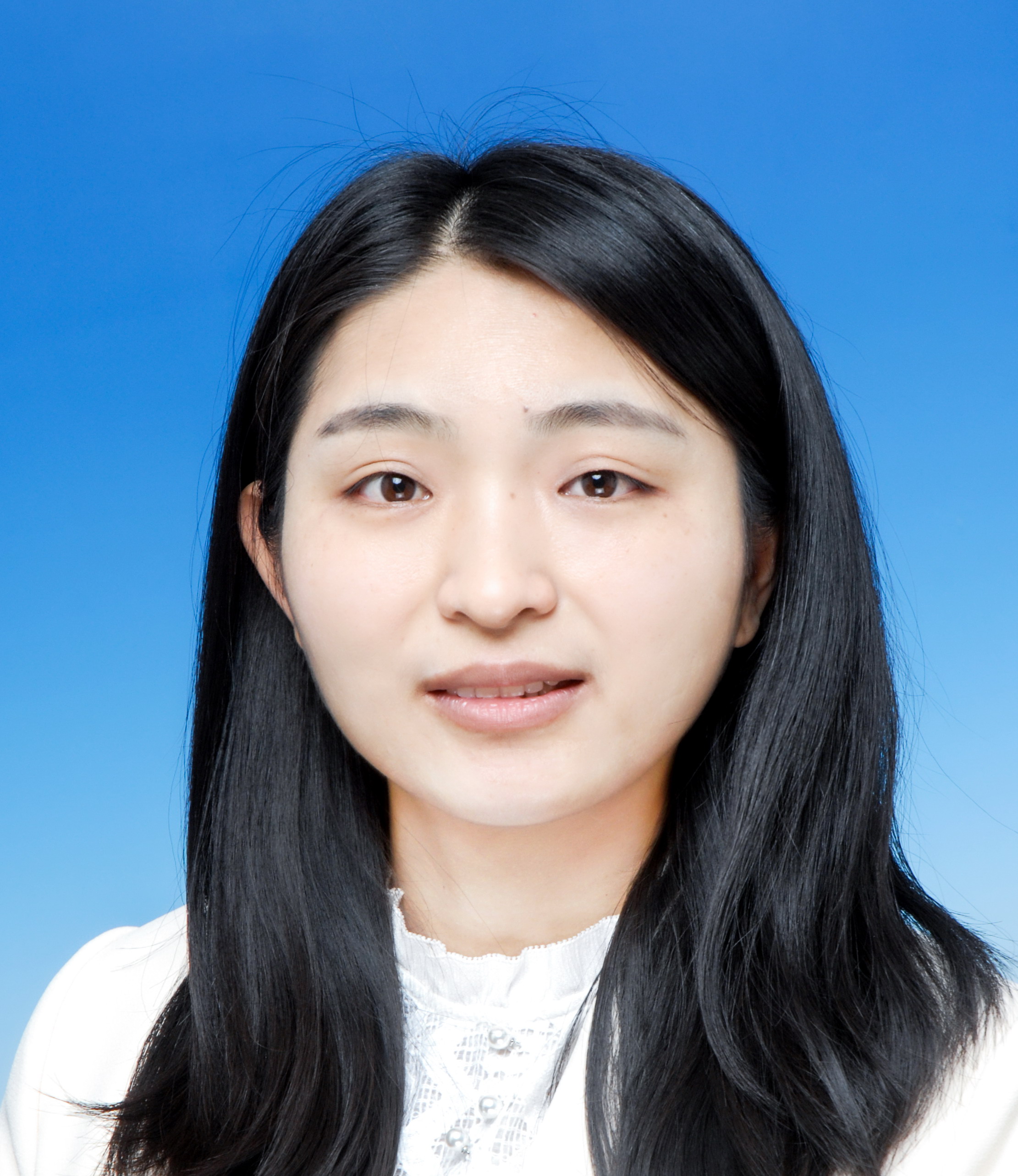}}]{Zhu Teng}
received her B.S. degree in automation from Central South University, Changsha, China, in 2006, and the Ph.D. degree in intelligent control and automation from Pusan National University, South Korea, 2013. From 2013 to 2015, she was a postdoctoral researcher with the School of Computer and Information Technology, Beijing Jiaotong University. She is now a professor in the School of Computer Science and Technology, Beijing Jiaotong University. Her current research interests are computer vision, machine learning, and deep learning.
\vspace{-30pt}
\end{IEEEbiography}

\begin{IEEEbiography}[{\includegraphics[width=1in,height=1.25in,clip,keepaspectratio]{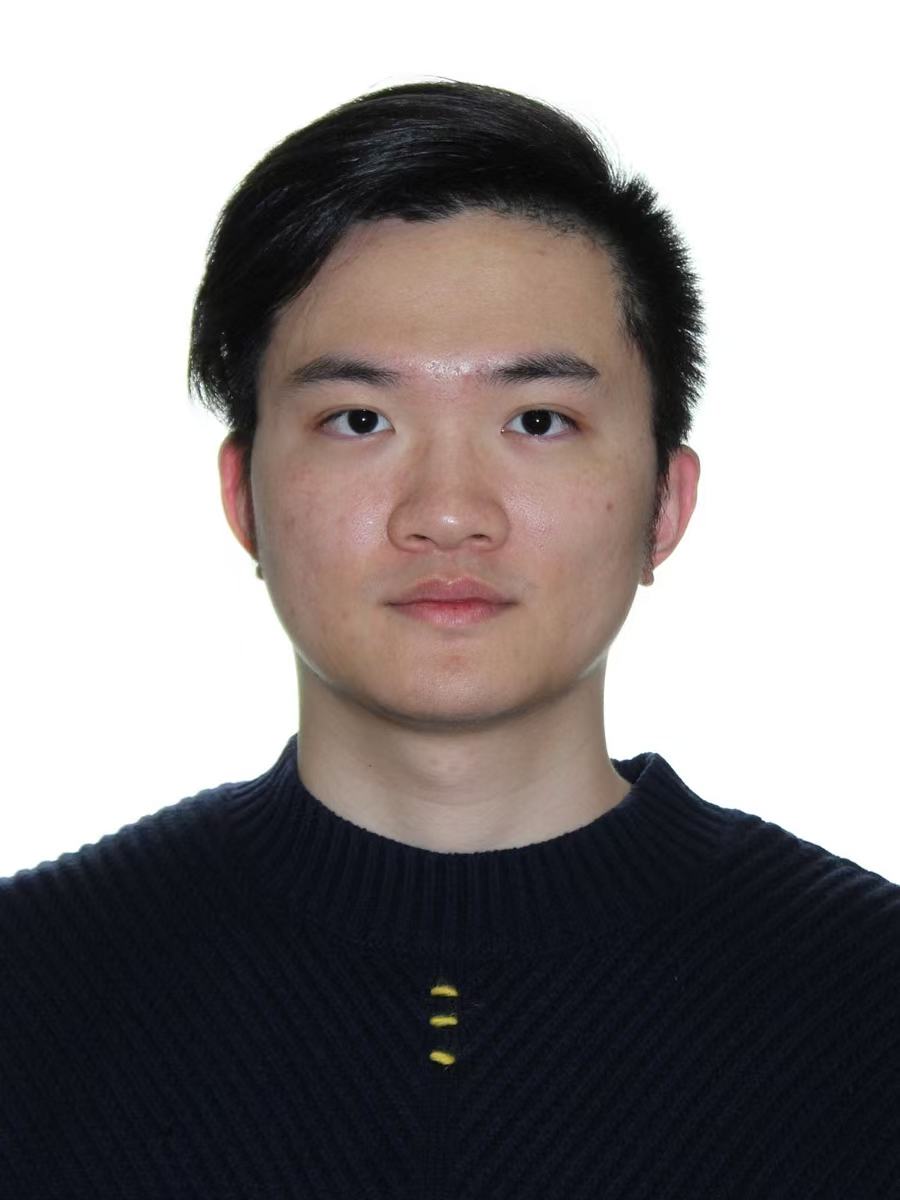}}]{Zhixiang Chi}
received his Ph.D. degree from the Edward S. Rogers Sr. Department of Electrical and Computer Engineering at the University of Toronto, Ontario, Canada, in 2025. He is currently a researcher in the Multimedia Laboratory within the ECE Department at the University of Toronto. His research interests include model adaptation and learning under distributional shift.
% \vspace{-10pt}
\end{IEEEbiography}

\begin{IEEEbiography}[{\includegraphics[width=1in,height=1.25in,clip,keepaspectratio]{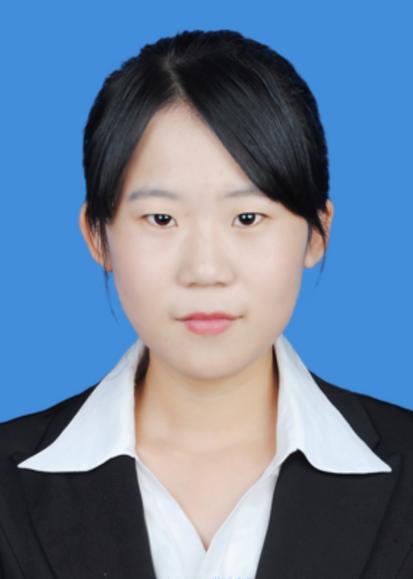}}]{Junnan Zou}
received the B.S. degree in Measurement and Control Technology and Instruments and the M.S. degree in Control Engineering from Yanshan University in 2017 and 2020, Qinhuangdao, China. She is currently pursuing the Ph.D. degree with the School of Computer Science and Technology, Beijing Jiaotong University. Her research interesteds include few shot learning, mechine learning, and computer vision.
\vspace{-25pt}
\end{IEEEbiography}

\begin{IEEEbiography}[{\includegraphics[width=1in,height=1.25in,clip,keepaspectratio]{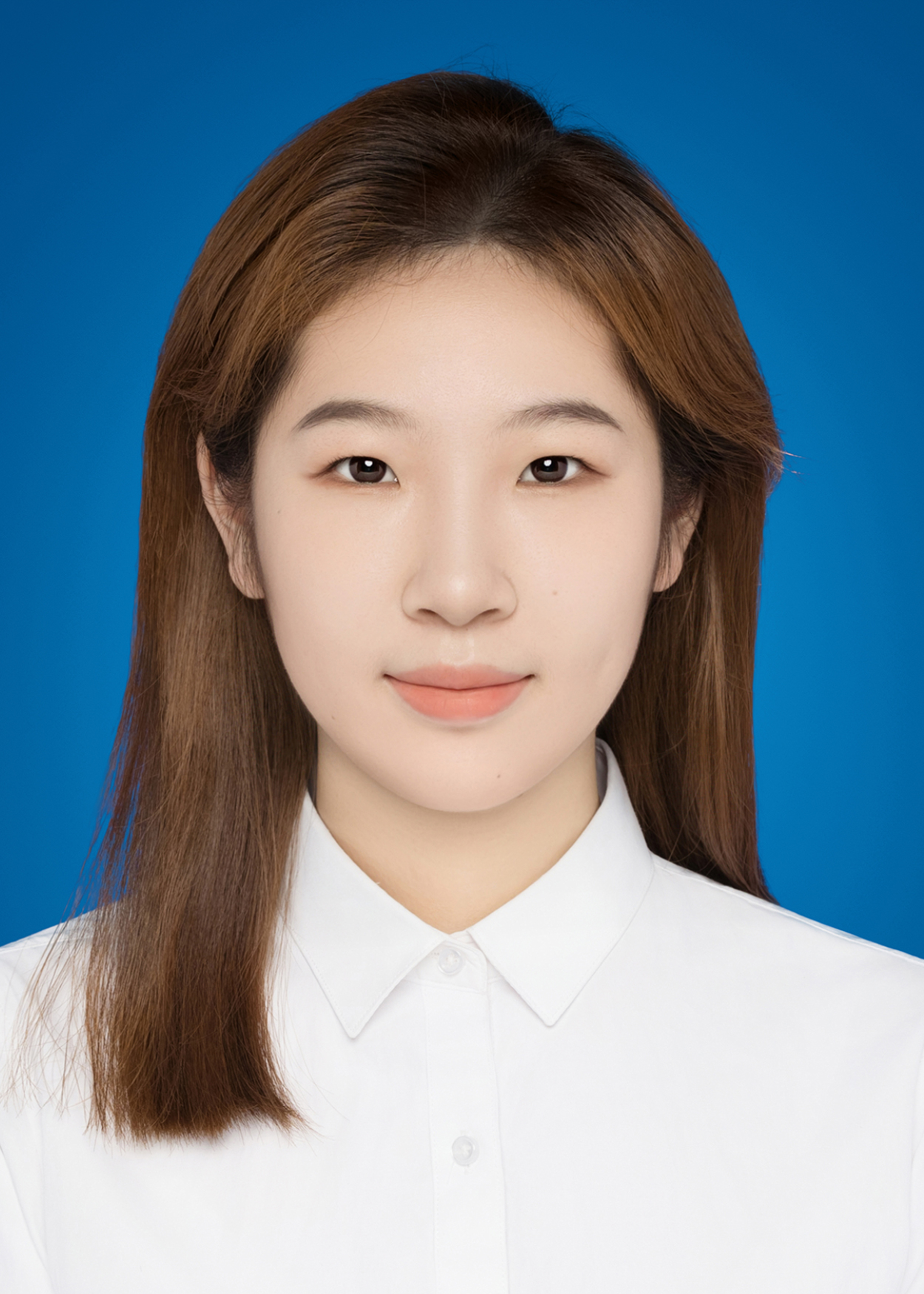}}]{Wenxin Luo}
received her B.S. degree in software engineering form Wuhan University of Technology, Wuhan, China in 2023. At present, she is studying for her master’s degree in computer science at Beijing Jiaotong University. Her current research interests include multimodal fake news detection, computer vision and deep learning.
\vspace{-25pt}
\end{IEEEbiography}

\begin{IEEEbiography}[{\includegraphics[width=1in,height=1.25in,clip,keepaspectratio]{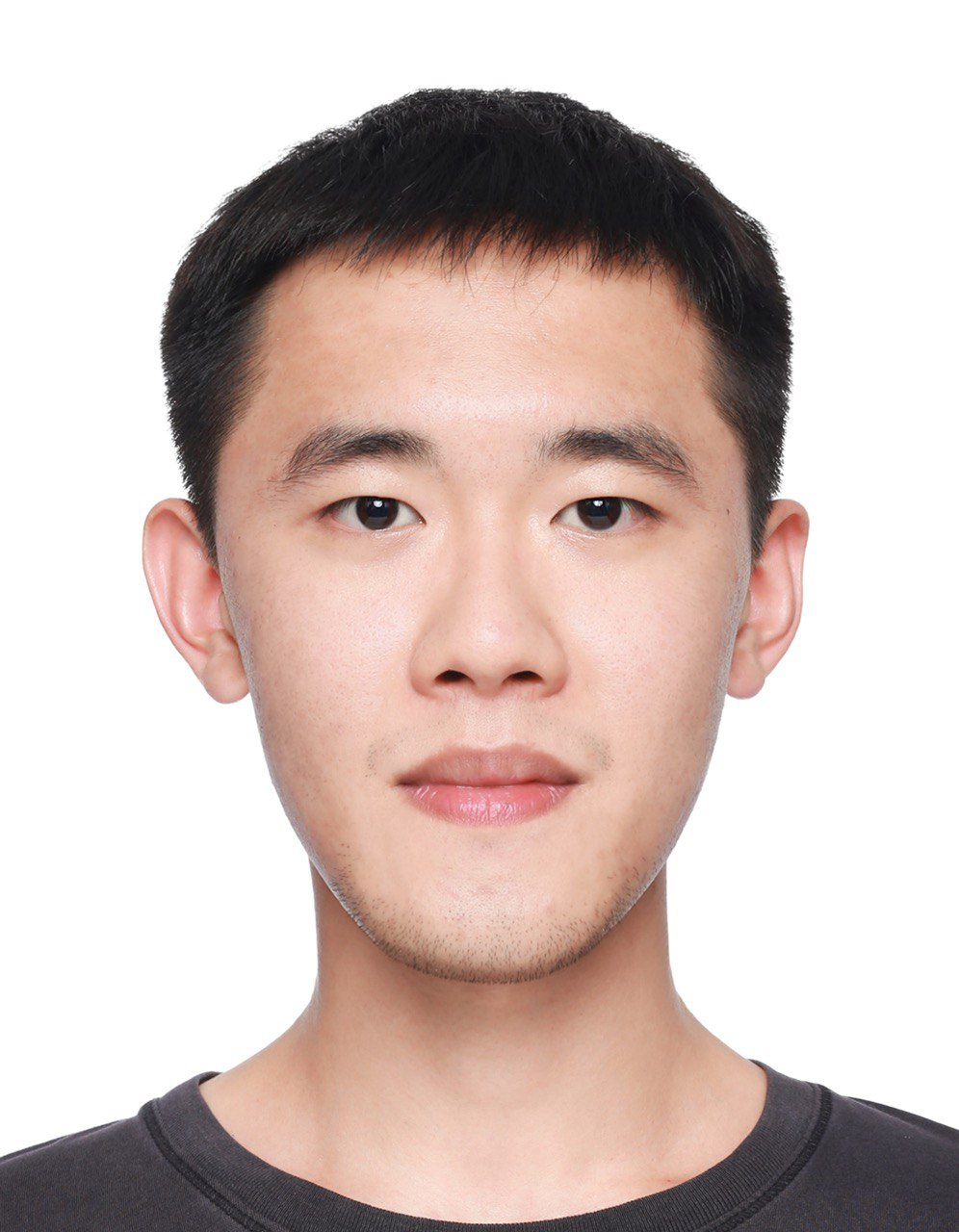}}]{Linfeng Ye}
is currently a Ph.D. student in the Edward S. Rogers Sr. Department of Electrical and Computer Engineering at the University of Toronto, Ontario, Canada, where he is a member of the Multimedia Laboratory. His research interests include information theory, computational pathology, and computer vision.
\vspace{-25pt}
\end{IEEEbiography}

\begin{IEEEbiography}[{\includegraphics[width=1in,height=1.25in,clip,keepaspectratio]{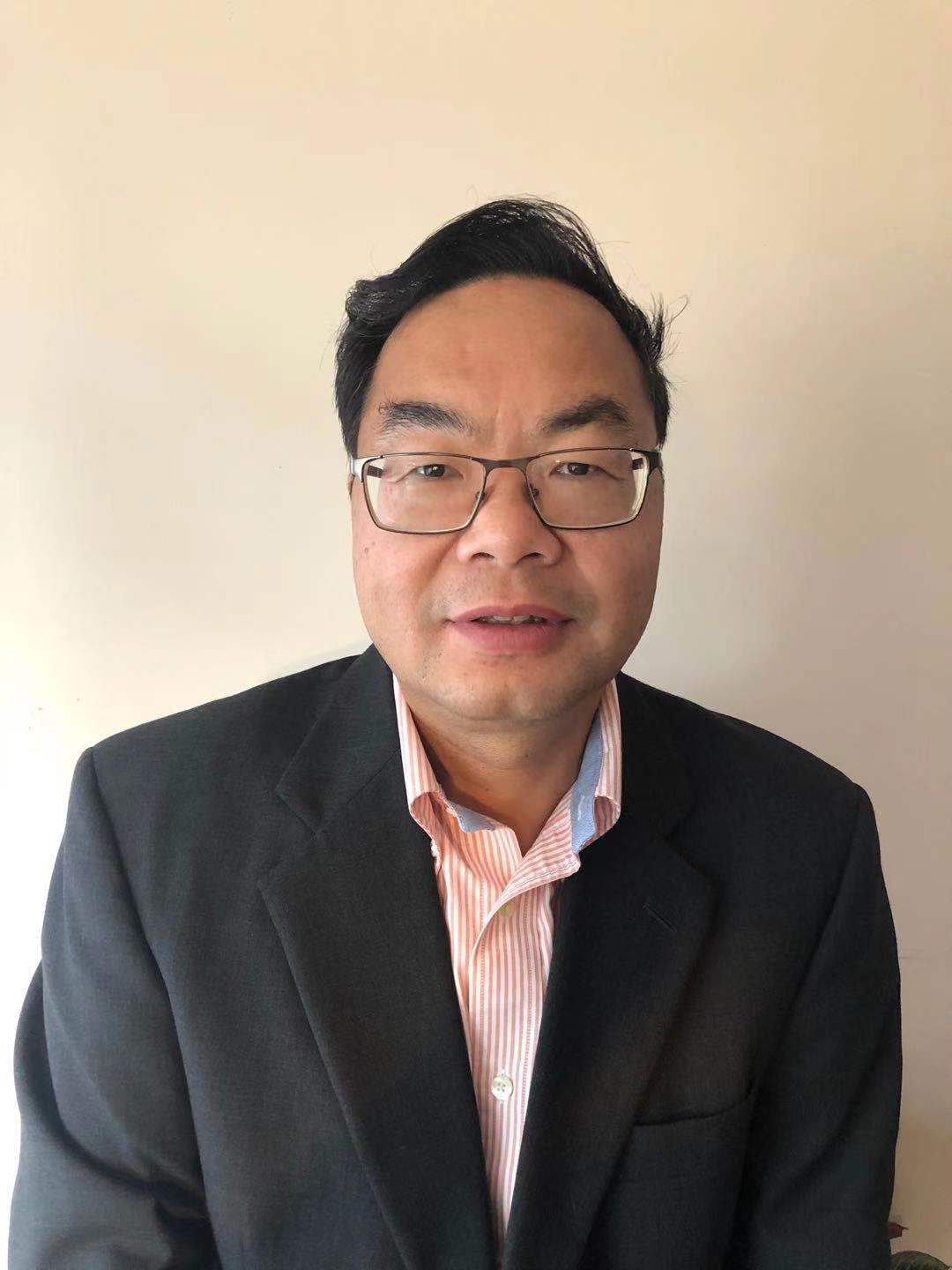}}]{Jianping Fan}
was a professor at UNC-Charlotte. He received his MS degree in theory physics from Northwestern University, Xian, China in 1994 and his PhD degree in optical storage and computer science from Shanghai Institute of Optics and Fine Mechanics, Chinese Academy of Sciences, Shanghai, China, in 1997. He was a Postdoc Researcher at Fudan University, Shanghai, China, during 1997-1998. From 1998 to 1999, he was a Researcher with Japan Society of Promotion of Science (JSPS), Department of Information System Engineering, Osaka University, Osaka, Japan. From 1999 to 2001, he was a Postdoc Researcher in the Department of Computer Science, Purdue University, West Lafayette, IN. His research interests include image/video privacy protection, automatic image/video understanding, and large-scale deep learning.
\vspace{-25pt}
\end{IEEEbiography}

\begin{IEEEbiography}[{\includegraphics[width=1in,height=1.25in,clip,keepaspectratio]{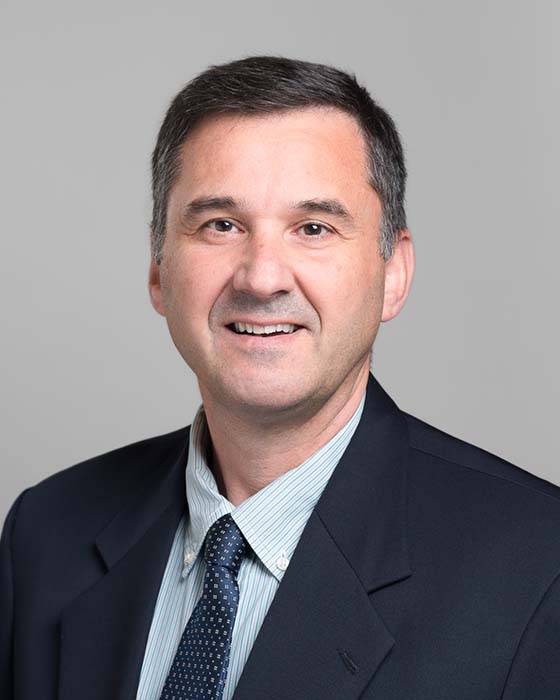}}]{Konstantions N. Plataniotis}
(Fellow, IEEE) received the B.Eng. degree in computer engineering from the University of Patras, Greece, and the M.S. and Ph.D. degrees in electrical engineering from the Florida Institute of Technology, Melbourne, FL, USA. He has been holding the Bell Canada Endowed Chair in Multimedia, since 2014. He is currently a Professor with The Edward S. Rogers Sr. Department of Electrical and Computer Engineering, University of Toronto, Toronto, ON, Canada, where he directs the Multimedia Laboratory. His research interests include image/signal processing, machine learning, adaptive learning systems, visual data analysis, multimedia and knowledge media, and affective computing. He is a fellow of the Engineering Institute of Canada and the Canadian Academy of Engineering. He was the Technical Co-Chair of the IEEE 2013 International Conference on Acoustics, Speech and Signal Processing. He served as the Inaugural IEEE Signal Processing Society Vice President for Membership (2014–2016) and the General Co-Chair for the 2017 IEEE GLOBALSIP. He served as the General Co-Chair for the 2018 IEEE International Conference on Image Processing (ICIP 2018) and the IEEE International Acoustics, Speech and Signal Processing (ICASSP 2021). He will be the General Chair for the 2027 IEEE International Conference on Acoustics, Speech and Signal Processing (ICASSP 2027), Toronto. He has served as the Editor-in-Chief for the IEEE SIGNAL PROCESSING LETTERS. He is a registered Professional Engineer in Ontario.
\end{IEEEbiography}

\vfill

\end{document}